%% file: main.tex
\documentclass[a4paper,fleqn]{cas-dc}
\usepackage{fontawesome5}
\usepackage{cite,chapterbib,placeins,microtype}
\usepackage{tabularx,fvextra,needspace}
\hypersetup{hidelinks,pdftitle={TriPhysBench: Evaluating Physical Intelligence for Embodied Agents}}
\DefineVerbatimEnvironment{verbatim}{Verbatim}{breaklines=true,breakanywhere=true,fontsize=\small,samepage=true,frame=single,framerule=0.3pt,framesep=5pt}
\providecommand{\tightlist}{\setlength{\itemsep}{0pt}\setlength{\parskip}{0pt}}
\pretocmd{\subsection}{\Needspace{4\baselineskip}}{}{}
\AtBeginEnvironment{thebibliography}{\footnotesize}
\apptocmd{\thebibliography}{\setlength{\itemsep}{1pt}\setlength{\parsep}{0pt}\setlength{\emergencystretch}{5em}}{}{}
\newcommand{\modelicon}[1]{\raisebox{-1.7pt}{\includegraphics[width=8.5pt,height=8.5pt,keepaspectratio]{figures/assets/model_icons/#1.png}}\hspace{2.5pt}}

\let\printorcid\relax
\ExplSyntaxOn
\cs_set:Npn \__first_foot:
{
  \parbox[t]{\textwidth}
  {\rule{\textwidth}{.2pt}\\ \sffamily\small TriPhysBench \hfill \thepage}
}
\cs_set_eq:NN \__cas_foot: \__first_foot:
\ExplSyntaxOff
\begin{document}
\shorttitle{TriPhysBench: Physical Intelligence for Embodied Agents}
\shortauthors{Sun et al.}
\title[mode=title]{TriPhysBench: Evaluating Physical Intelligence for Embodied Agents}

\author[1]{Zhongxiang Sun}
\ead{sunzhongxiang@ruc.edu.cn}
\author[2]{Chenghao Yang}
\ead{chyang@stu.zuel.edu.cn}
\author[3]{Haojie Ding}
\ead{dinghaojie@kuaishou.com}
\author[3]{Boheng Zhang}
\ead{yangxiao16@kuaishou.com}
\author[3]{Fan Yang}
\ead{yangfan@kuaishou.com}
\author[4]{Qingsong Wen}
\ead{qingsongedu@gmail.com}
\author[5]{Kun Wang}
\ead{wang.kun@ntu.edu.sg}

\affiliation[1]{organization={Gaoling School of Artificial Intelligence, Renmin University of China},city={Beijing},country={China}}
\affiliation[2]{organization={School of Information Engineering, Zhongnan University of Economics and Law},city={Wuhan},country={China}}
\affiliation[3]{organization={Kuaishou Technology},city={Beijing},country={China}}
\affiliation[4]{organization={Squirrel AI Learning}}
\affiliation[5]{organization={School of Artificial Intelligence and Automation, Huazhong University of Science and Technology},city={Wuhan},country={China}}
\nonumnote{Qingsong Wen: Senior Member, IEEE.}

\begin{abstract}
Embodied agents need physical intelligence to understand their surroundings through physical perception, physical knowledge retrieval, and physical reasoning. These capabilities must work together to interpret events and make quantitative judgments. Existing benchmarks advance physical understanding and general-purpose tool use, yet the coordination of these capabilities in physical scenarios remains insufficiently characterized. Inspired by Popper's Three Worlds, we introduce TriPhysBench, a benchmark of 410 questions from 200 videos that brings physical perception, physical knowledge retrieval, and physical reasoning into a shared evaluation setting. Its measurement-grounded and knowledge-grounded subtasks impose complementary demands on obtaining observations and acquiring external specifications, then applying them to quantitative problems. Among six representative vision-language models, the best achieves an answer accuracy of only 50.38\%, revealing limited performance on tasks that require these capabilities to work together. We further introduce TriPhys, a framework that learns a unified policy for physical perception, knowledge retrieval, and quantitative reasoning from complete tool-integrated trajectories. The policy uses intermediate reasoning to guide retrieval and computation, then incorporates tool feedback into subsequent physical inference. On 110 questions from videos excluded from training, TriPhys improves answer and solution scores across three Qwen configurations, with a 17.7\% relative gain in answer accuracy for TriPhys-27B over its Qwen3.6-27B backbone. This work provides a benchmark and a learning pathway toward embodied agents that move beyond recognizing events to understanding and reasoning about the physical processes behind them.
\end{abstract}

\begin{keywords}
Embodied intelligence \sep physical perception \sep physical knowledge retrieval \sep physical reasoning \sep multimodal agents
\par\medskip
\noindent\faGithub\ \href{https://github.com/yangch0429/TriPhys}{\textbf{Code \& Data}}\par
\smallskip\noindent\faGlobe\ \href{https://yangch0429.github.io/TriPhys/}{\textbf{Project Website}}\par
\end{keywords}

\include{TriPhysBench_main}

\include{TriPhysBench_appendix}
\end{document}

%% file: TriPhysBench_main.tex
\maketitle
\hypersetup{pdfauthor={Zhongxiang Sun, Chenghao Yang, Haojie Ding, Boheng Zhang, Fan Yang, Qingsong Wen, Kun Wang}}
\makeatletter\immediate\write\@auxout{\string\citation{BSTcontrol}}\makeatother

\section{Introduction}\label{sec:introduction}

\begin{figure*}[pos=!t]
\centering
\includegraphics[width=0.94\textwidth]{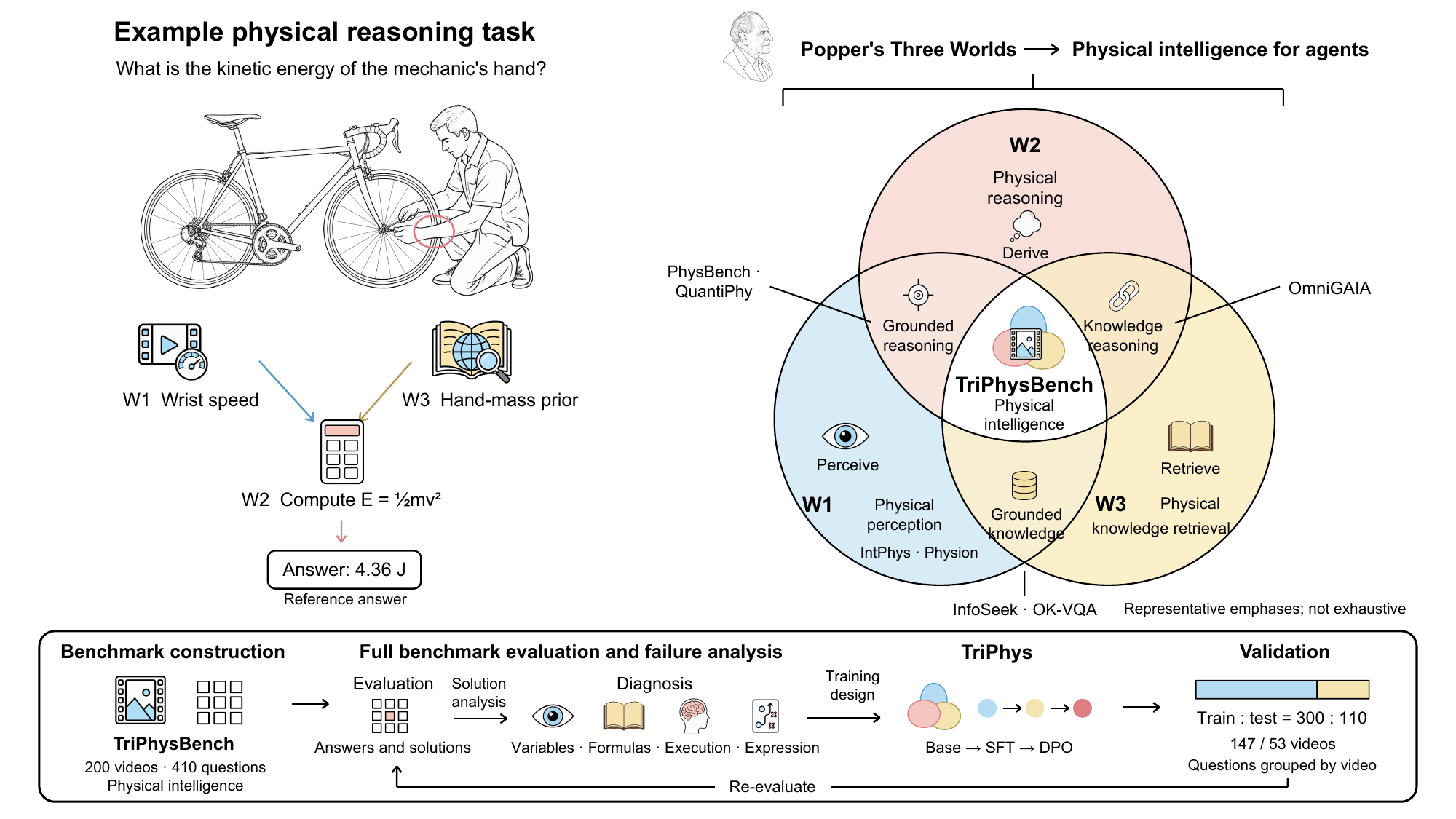}
\caption{Overview of TriPhysBench and TriPhys. Blue, yellow, and pink denote physical perception (W1), knowledge retrieval (W3), and reasoning (W2), respectively. Scenes and workflow are schematic.}
\label{fig:overview}
\end{figure*}

Embodied agents need to connect observations and knowledge with the physical conditions of their tasks \cite{driess2023palme,ahn2023saycan,brohan2023rt2,li2024manipllm,ji2025robobrain,li2024embodied}. This requires interpreting observations, obtaining relevant properties, and applying physical laws \cite{wu2015galileo,chow2025physbench,zhai2024physical}. We study physical intelligence as the coordination of physical perception, physical knowledge retrieval, and physical reasoning. To characterize the complementary roles of these capabilities, we draw on Popper's Three Worlds \cite{popper1978three}, associating perception with the physical world (W1), retrieval with objective knowledge (W3), and reasoning with mental activity (W2). In the schematic example in Figure \ref{fig:overview}, physical perception estimates wrist speed from the video, knowledge retrieval supplies a hand-mass prior, and physical reasoning combines them to estimate kinetic energy. Accurate physical reasoning requires the quantities recovered from video, the retrieved properties, and the modeling assumptions to be consistent with the same physical situation.

Physical understanding benchmarks test plausibility and event prediction \cite{riochet2018intphys,bear2021physion,yi2020clevrer}, while PhysBench broadens coverage across properties, relationships, scenes, and dynamics \cite{chow2025physbench}. QuantiPhy advances quantitative video reasoning by estimating kinematic quantities using a supplied physical prior \cite{li2025quantiphy}. Multidisciplinary benchmarks such as MMMU, OlympiadBench, and MMVU test expert-level scientific reasoning over images and videos \cite{yue2024mmmu,he2024olympiadbench,zhao2025mmvu}. Knowledge-based visual benchmarks examine access to external information \cite{marino2019okvqa,chen2023infoseek}, and GAIA and OmniGAIA evaluate general-purpose agents on reasoning and tool use across information sources and modalities \cite{mialon2024gaia,li2026omnigaia}. EmbodiedBench further tests vision-driven agents across high-level planning and low-level navigation and manipulation \cite{yang2025embodiedbench}. Across these settings, a central question remains insufficiently examined: how reliably can agents acquire the required physical quantities, select knowledge applicable to the observed conditions, and use both in quantitative reasoning?

We introduce TriPhysBench, comprising 410 questions from 200 videos of physical events and human activities. To construct the benchmark, we adapt event-graph-based question generation \cite{li2026omnigaia} to connect video observations, external properties, and physical relations through physical event graphs (Figure \ref{fig:construction}). The resulting questions form two complementary subtasks. Measurement-grounded Physical Reasoning (MPR, 228 questions) uses supported video measurements and requires solvers to recover designated observations from the video when they are withheld from the question. Knowledge-grounded Physical Reasoning (KPR, 182 questions) provides the video context and stated observations, while requiring the retrieval, selection, and application of external physical specifications. Across both subtasks, withholding selected observations or specifications controls difficulty, and questions take four answer forms: point estimation, interval estimation, physical judgment, and scenario comparison. Large language model review and rule-based checks screen the questions and their reference solutions.

Our evaluation on TriPhysBench reveals recurring errors in identifying relevant physical quantities and carrying them through numerical derivations, highlighting the need to coordinate information acquisition with physical reasoning. We therefore propose TriPhys, a tool-integrated framework that learns a unified policy for physical perception, knowledge retrieval, and quantitative reasoning. Intermediate reasoning guides retrieval and computation, while tool feedback informs subsequent physical inference. The policy is trained on complete tool-integrated trajectories and refined through preference optimization \cite{gou2024tora,qin2024toolllm,rafailov2023dpo}.

We evaluate six vision-language models (VLMs): GPT-5.5 \cite{openai2026gpt55}, Gemini-3.6-Flash \cite{google2026gemini36flash}, Qwen3.6-27B \cite{qwen2026qwen36_27b}, Qwen3.5-9B \cite{qwen2026qwen35_9b}, Qwen3.5-4B \cite{qwen2026qwen35_4b}, and InteractiveOmni-4B \cite{sensenova2025interactive}. On the full benchmark, even the strongest model, Gemini-3.6-Flash, achieves only 50.38\% answer accuracy, exposing a substantial gap in coordinating physical information and reasoning. On held-out videos, TriPhys improves answer and solution scores across all three Qwen configurations, with relative accuracy gains of 16.4\%--17.7\%. TriPhys-27B, adapted from Qwen3.6-27B, reaches 58.18\% accuracy, exceeding GPT-5.5's 51.14\% on the same test set. Diagnosis of three models identifies weaknesses in variable identification and conclusion formation.
Together, TriPhysBench and TriPhys provide a foundation for evaluating and developing physical intelligence in embodied agents.

Our main contributions are:

\begin{itemize}
\tightlist
\item
  We introduce \textbf{TriPhysBench}, a benchmark with two complementary subtasks constructed through physical event graphs and controlled information withholding to evaluate the coordination of physical perception, knowledge retrieval, and reasoning.
\item
  We propose \textbf{TriPhys}, a framework that combines supervision of complete tool-integrated trajectories with preference optimization for physical reasoning, improving answer and solution scores across three Qwen configurations on videos excluded from training.
\item
  We provide \textbf{systematic evaluations and analyses} of six VLMs across both subtasks, including failure diagnosis of three models that examines variable identification, relation selection, numerical derivation, and conclusion formation.
\end{itemize}

\Needspace{8\baselineskip}

\section{Related Work}\label{sec:related-work}

\begin{table*}[pos=!t]
\caption{Comparison with representative benchmarks.}
\label{tab:benchmark-comparison}
\centering
\includegraphics[width=\textwidth]{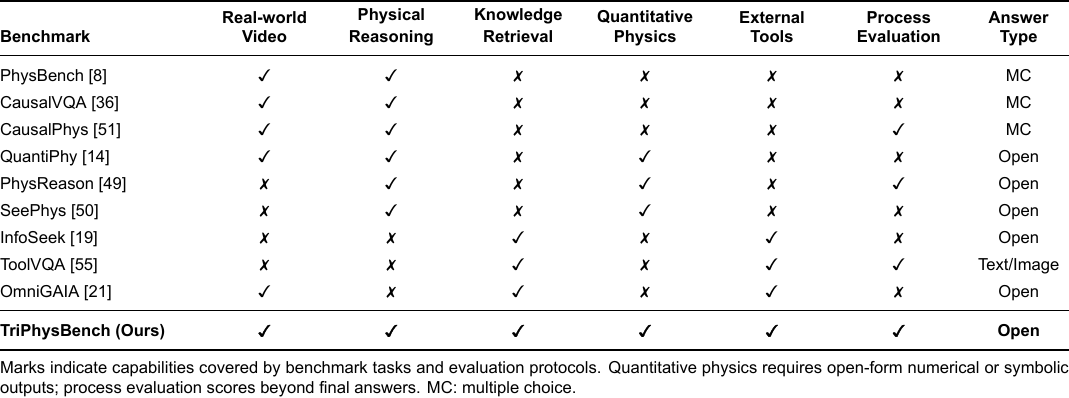}
\end{table*}

\subsection{Physical Intelligence Evaluation}\label{video-understanding-and-physical-reasoning}

Physical evaluation spans several task formats. PIQA tests physical commonsense in language \cite{bisk2020piqa}, while PHYRE requires actions that solve simulated physics puzzles \cite{bakhtin2019phyre}. Controlled-scene benchmarks assess plausibility, interaction prediction, and causal events \cite{riochet2018intphys,bear2021physion,yi2020clevrer}; IntPhys 2 extends possible-versus-impossible event discrimination to more complex synthetic environments \cite{bordes2025intphys2}. PhysBench covers object properties, relationships, scenes, and dynamics \cite{chow2025physbench}, whereas PhysGame examines physical violations in gameplay videos \cite{cao2024physgame}. In real videos, CausalVQA tests causal understanding through anticipation, planning, and hypothetical or counterfactual questions \cite{foss2025causalvqa}, and QuantiPhy estimates object size, velocity, and acceleration using a given physical prior \cite{li2025quantiphy}. Spatial reasoning benchmarks and models probe distances, sizes, and spatial relations in images and videos \cite{chen2024spatialvlm,cheng2024spatialrgpt,yang2025thinking}, while EgoThink and OpenEQA assess understanding from an agent's first-person viewpoint \cite{cheng2024egothink,majumdar2024openeqa}. Generative interactive environments and video generation models raise the complementary question of whether learned dynamics follow physical laws \cite{bruce2024genie,kang2025howfar}. These benchmarks examine distinct aspects of physical competence, from judging event plausibility to estimating numerical quantities.

Scientific problem-solving benchmarks such as SciBench and OlympicArena evaluate multistep reasoning across disciplines \cite{wang2024scibench,huang2024olympicarena}, and general multimodal evaluations examine capability coverage, reliance on visual content, and visual hallucination \cite{li2024seedbench,chen2024mmstar,guan2024hallusionbench}. Beyond final answers, PhysReason evaluates multistep physics solutions at answer and step levels \cite{zhang2025physreason}, and SeePhys emphasizes extracting information from physics diagrams \cite{xiang2025seephys}. CausalPhys uses annotated dependency graphs for reasoning assessment and causal rationale supervision \cite{tang2026causalphys}. TriPhysBench evaluates physical perception, physical knowledge retrieval, and physical reasoning jointly within each task. Its measurement-grounded and knowledge-grounded subtasks place complementary demands on recovering observations and selecting external physical specifications. Both require relating this information to the scene and applying it to quantitative estimation, physical judgment, or scenario comparison. Performance under these coordinated demands is evaluated through the returned answers and solutions.

\subsection{Knowledge Retrieval and Tool Use}\label{external-knowledge-and-tool-use}

Knowledge-based visual question answering connects images with information beyond their visible content. OK-VQA and A-OKVQA require outside knowledge and commonsense \cite{marino2019okvqa,schwenk2022aokvqa}, while InfoSeek targets entity-specific information \cite{chen2023infoseek}. MuRAG retrieves from image--text memory for open question answering \cite{chen2022murag}; ReflectiVA learns to assess the need for retrieval and the relevance of retrieved knowledge \cite{cocchi2025reflectiva}. ToolVQA evaluates multistep visual reasoning with external tools \cite{yin2025toolvqa}, and MMSearch-R1 learns image and text search policies through reinforcement learning \cite{wu2026mmsearch}. Agent benchmarks such as VisualWebArena, GTA, and TravelPlanner evaluate multimodal web interaction, real-world tool invocation, and constrained planning \cite{koh2024visualwebarena,wang2024gta,xie2024travelplanner}; CodeAct uses executable code as a unified action space \cite{wang2024codeact}, and WebThinker equips reasoning models with autonomous web search for deep research \cite{li2025webthinker}. GAIA and OmniGAIA extend evaluation to reasoning and multi-turn tool execution across information sources and modalities \cite{mialon2024gaia,li2026omnigaia}. Our focus is the physical applicability of acquired information: a retrieved property must match the observed object and conditions before entering a calculation. This connects knowledge acquisition with perception and quantitative reasoning in the same evaluation (Table~\ref{tab:benchmark-comparison}).

\subsection{Learning Tool-Integrated Reasoning}\label{trajectory-supervision-and-preference-learning}

ReAct interleaves reasoning and actions to gather information and update plans \cite{yao2023react}; Toolformer learns when to call APIs and incorporate their results \cite{schick2023toolformer}. PAL delegates computation to executable programs \cite{gao2023pal}, while ToRA and ToolLLM learn from tool-integrated solutions and tool-use trajectories \cite{gou2024tora,qin2024toolllm}. Direct Preference Optimization (DPO) learns from preferred and rejected outputs \cite{rafailov2023dpo}. Subsequent variants remove the reference model \cite{meng2024simpo}, iterate preference optimization over reasoning chains \cite{pang2024iterative}, or improve a model against its own earlier outputs \cite{chen2024spin}, and ETO constructs preference pairs from successful and failed agent trajectories \cite{song2024trial}. In multimodal models, RLHF-V aligns behavior using fine-grained correctional feedback \cite{yu2024rlhfv}; Visual CoT, compositional chain-of-thought prompting, LLaVA-CoT, and Insight-V elicit structured or long-chain visual reasoning \cite{shao2024visualcot,mitra2024ccot,xu2025llavacot,dong2025insightv}; and Visual-RFT and Video-R1 apply reinforcement fine-tuning with rule-based rewards to images and videos \cite{liu2025visualrft,feng2025videor1}. For multimodal tool use, VITAL combines trajectory supervision and reinforcement learning for on-demand video perception \cite{zhang2026thinking}, and VTool-R1 learns Python-based visual tool use from outcome rewards \cite{wu2026vtool}. In physical reasoning, PhysGame introduces instruction and preference data for training PhysVLM \cite{cao2024physgame}, while IPR-1 learns through world-model rollouts in interactive games \cite{zhang2026ipr}. OmniAtlas combines tool-integrated trajectory supervision with preference optimization for omni-modal tasks \cite{li2026omnigaia}. TriPhys applies these learning principles to complete physical solutions, preserving dependencies between observations, retrieved properties, and numerical derivations. Intermediate reasoning guides tool calls, and tool feedback informs subsequent inference.

\section{TriPhysBench Construction}\label{sec:construction}

\begin{figure*}[pos=!t]
\centering
\includegraphics[width=\textwidth]{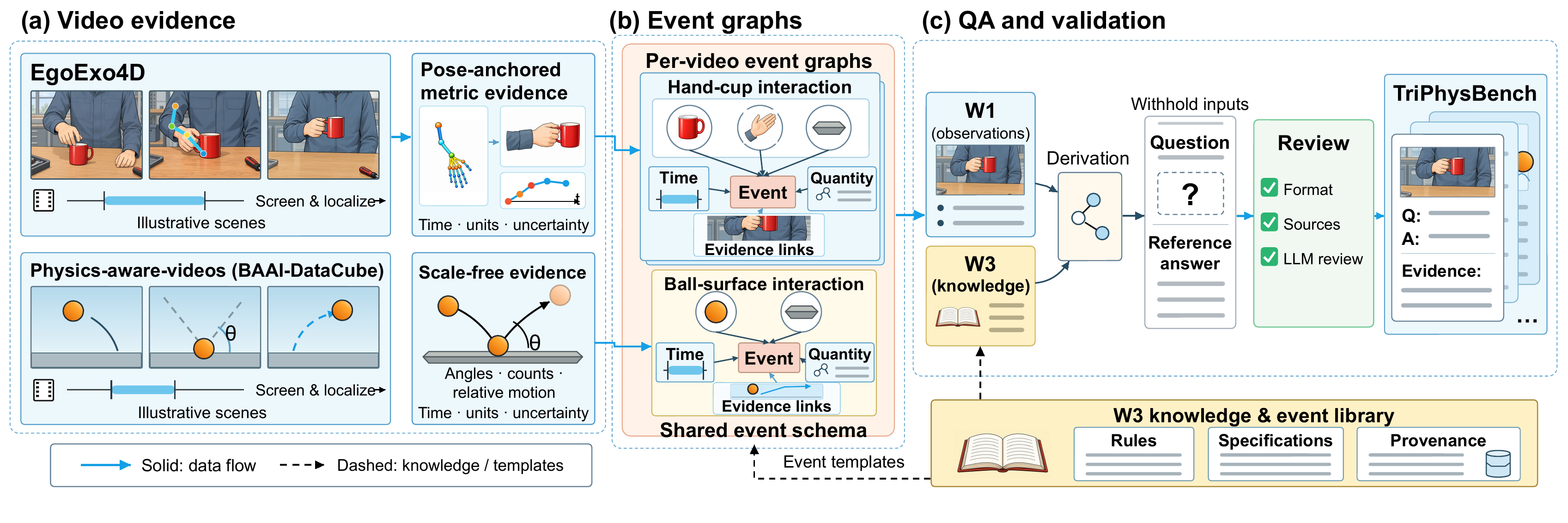}
\caption{TriPhysBench construction: (a) video processing for MPR (EgoExo4D) and KPR (Physics-aware-videos); (b) physical event graphs; (c) question generation and review. Solid arrows indicate data flow; dashed arrows indicate external inputs. Scenes are schematic.}
\label{fig:construction}
\end{figure*}

We construct TriPhysBench to evaluate how embodied agents coordinate physical perception, physical knowledge retrieval, and physical reasoning when solving problems about real-world events. The benchmark comprises 410 questions from 200 videos, organized into complementary measurement-grounded and knowledge-grounded subtasks. By varying the observations and knowledge provided in each question, these subtasks examine how agents acquire and apply the physical information needed for quantitative estimation and judgment. Figure \ref{fig:construction} summarizes the construction pipeline.

\subsection{Task Design and Video Sources}\label{sec:task}\label{sec:sources}

Each question pairs a video with a physical problem and asks for a supported solution and final answer. The solver receives any observations explicitly stated in the question and can use external retrieval and numerical computation. Physical perception supplies relevant observations, knowledge retrieval supplies applicable properties or relations, and reasoning combines them into an answer. Reference solutions and supporting records are retained as construction and evaluation annotations.

\textbf{Measurement-grounded Physical Reasoning.} This subtask contains 228 questions from 90 EgoExo4D recordings of skilled activities, including bicycle repair and sports \cite{grauman2024egoexo4d}. Construction uses a third-person view and three-dimensional body-pose annotations to establish supported measurements. Questions vary how many observations are given, requiring the solver to recover designated quantities from the video when they are withheld. The task combines this acquisition of observations with physical knowledge and quantitative reasoning.

\textbf{Knowledge-grounded Physical Reasoning.} This subtask contains 182 questions from 110 recordings in BAAI-DataCube's Physics-aware-videos collection \cite{baai2025physicsaware}, covering processes such as collisions, oscillations, fluid motion, and deformation. These videos lack the calibrated pose reference available in EgoExo4D. Questions therefore state the estimated observations and require the solver to obtain and apply external specifications, such as an object's mass or dimensions. The challenge is to select properties that match the physical situation and use them in the derivation.

Both subtasks use four answer forms: point estimation, interval estimation, physical judgment, and scenario comparison. These specify the requested output independently of the source or difficulty. The shared forms allow both acquisition settings to test numerical estimates as well as judgments and comparisons grounded in physical quantities. Appendix~\ref{app:construction} gives the corresponding TYPE labels and reference-answer conventions.

\begin{figure*}[pos=!t]
\centering
\includegraphics[width=\textwidth]{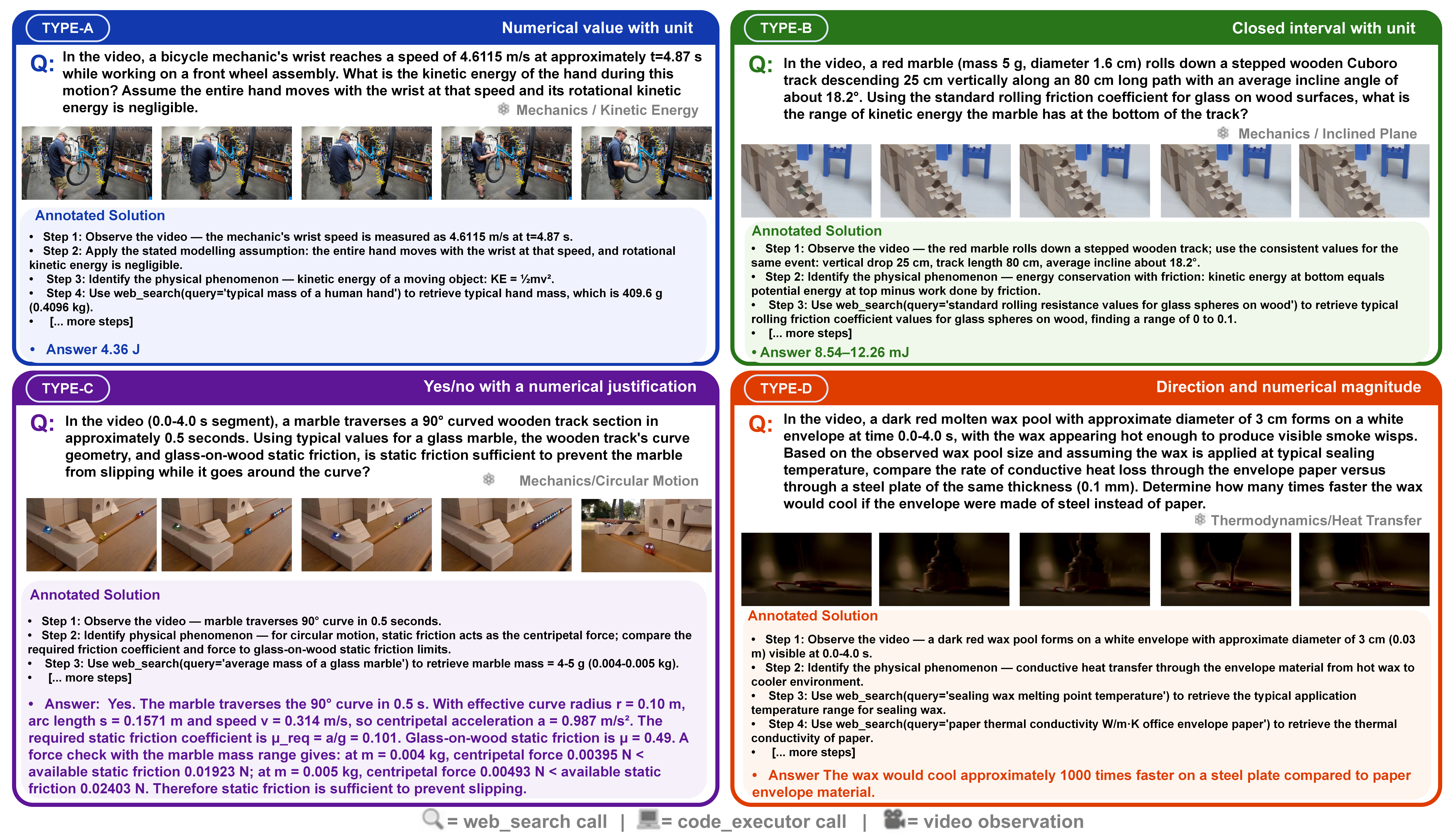}
\caption{Four answer forms illustrated by reference cases. Type A is an MPR case from EgoExo4D; Type B, C, and D are KPR cases from Physics-aware-videos. Highlighted frames mark key observations; the hand mass, friction coefficients, curve radius, and thermal conductivities follow the reference assumptions.}
\label{fig:question-examples}
\end{figure*}

The four cases in Figure \ref{fig:question-examples} illustrate how these requirements appear in individual questions. Video frames establish the event context, physical properties supply quantities needed for the calculation, and the reference solution connects them to the requested answer form. The examples cover hand motion during bicycle repair, a marble on a stepped track, a marble on a curved track, and heat conduction from molten sealing wax.

Video screening first matches source action descriptions or captions against physical event categories and then checks technical quality, visible motion, and whether an identifiable subject supports a quantitative question. Gemini-3.6-Flash \cite{google2026gemini36flash} then parses accepted recordings into intervals with scene and event descriptions that support observation extraction.

\subsection{Observation and Knowledge Acquisition}\label{sec:observation}

For the measurement-grounded subtask, pose-based human measurements and tracked object trajectories supply quantities tied to video intervals and available geometry. Physical distances and speeds require compatible coordinates and timestamps. Retained measurements include their supporting records and construction tolerances, establishing which observations can support a metric answer. These tolerances express the uncertainty adopted during construction. Appendix~\ref{app:construction} details measurement eligibility, extraction tools, and temporal alignment.

For the knowledge-grounded subtask, visual estimation targets event counts, angles, and relative quantities that do not require a pixel-to-meter conversion. Their interpretation still depends on the viewpoint and comparison reference. A separate retrieval step obtains mass, dimension, or material specifications for identifiable objects. A retrieved property must match the object's identity and the conditions of the question before it enters the reference derivation. Any additional absolute value explicitly stated in a question serves as a given problem condition. Together, the two routes provide observations and knowledge with defined physical roles.

\subsection{Event Representation and Question Generation}\label{sec:question-generation}

A physical event graph connects the participants and processes in each video to their observations, external properties, and physical relations. Event templates specify the information needed to derive a target quantity. DeepSeek-V3.2 \cite{deepseek2025v32} uses these templates to assemble candidate derivations and identify missing dependencies. Graph relations are supported by observations or applicable knowledge. This representation organizes question construction and reference solutions, linking the information to be acquired with its use in the physical calculation. The reference derivation records how the available observations and properties determine the target, with explicit units, target entities, and relevant event intervals. Quantities already stated in a question can be used directly; the hidden information defines what the solver must recover.

We generate questions by selecting a target and withholding quantities needed to recover it. In the measurement-grounded subtask, Easy, Medium, and Hard questions require recovery of zero, one, or two designated observations, subject to measurement availability. The 146 Medium and Hard MPR questions therefore require the solver to recover the omitted observations from the video. In the knowledge-grounded subtask, observations are stated and one, two, or three external specification values are withheld. An object can remain identifiable while its required numerical property is hidden. Difficulty thus varies the information that must be acquired before reasoning can be completed. Both routes retain the original physical event context.

\begin{figure*}[pos=!t]
\centering
\includegraphics[width=\textwidth]{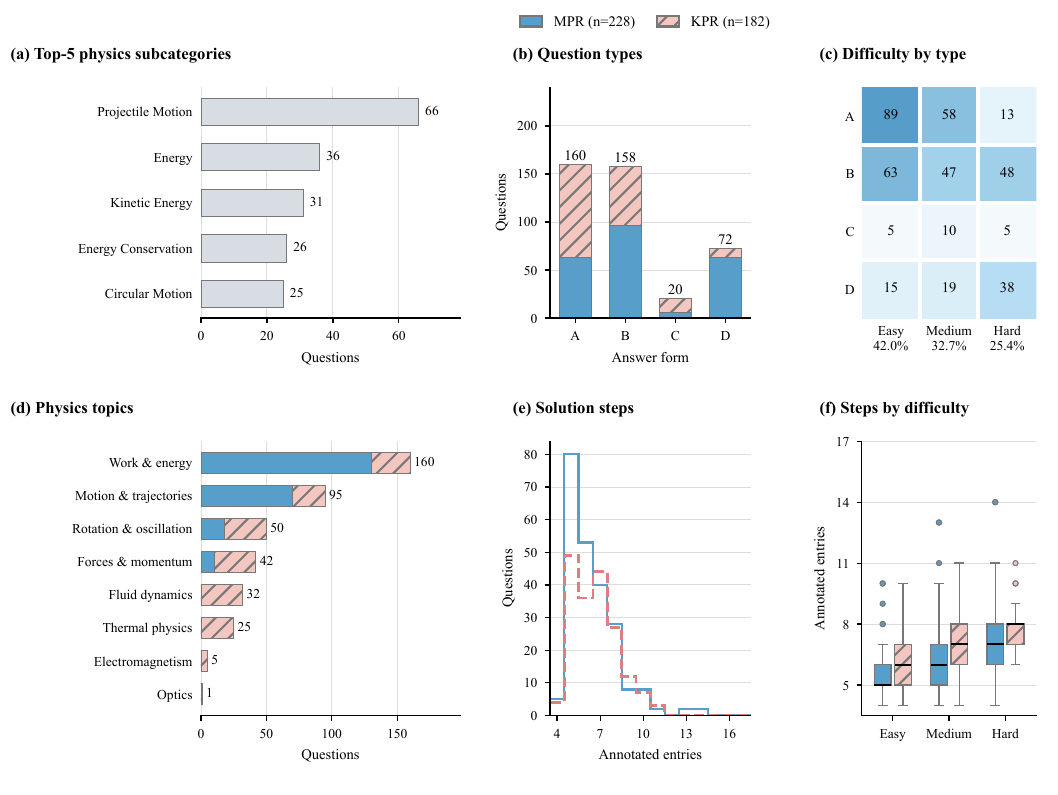}
\caption{TriPhysBench dataset statistics. A--D denote point estimation, interval estimation, physical judgment, and scenario comparison. Panel (d) groups all questions into eight physics topics. Boxplots show medians and interquartile ranges.}
\label{fig:statistics}
\end{figure*}

\subsection{Quality Review}\label{sec:quality-review}

Quality review assesses question clarity, answer correctness, support for physical quantities, and coherence of the reference solution. Large language model (LLM) review is followed by rule-based checks, and candidates that fail the acceptance criteria are excluded. In an additional review of all 410 questions, GPT-5.5 \cite{openai2026gpt55} and Gemini-3.6-Flash independently judge question understandability and answer correctness; together with the DeepSeek-V3.2 construction review, the three judges agree on both criteria for 95.6\% of assessed questions. Appendix~\ref{app:construction} specifies the assessment scope, raw agreement statistics, and detailed review procedure.

\subsection{Dataset Statistics}\label{sec:composition}

Aggregate performance on TriPhysBench is driven primarily by quantitative estimation, which accounts for over three quarters of its questions (Figure \ref{fig:statistics}b). The benchmark therefore places particular weight on converting observations and physical knowledge into consistent numerical answers. Work and energy form the largest shared topic, while the knowledge-grounded subtask extends coverage to fluid, thermal, electromagnetic, and optical phenomena (Figure \ref{fig:statistics}d). Overall scores therefore emphasize this shared mechanical core, while the less represented physical domains contribute comparatively little. Appendix~\ref{app:physics-topics} gives the topic mapping.

\begin{table}[pos=!t]
\caption{Dataset composition by subtask. Subcategory totals count distinct labels in the union.}
\label{tab:composition}
\centering\footnotesize
\setlength{\tabcolsep}{4pt}
\renewcommand{\arraystretch}{1.2}
\begin{tabularx}{\columnwidth}{@{}lXrrr@{}}
\toprule
\textbf{Category} & \textbf{Statistic} & \textbf{MPR} & \textbf{KPR} & \textbf{Total} \\
\midrule
 & Videos & 90 & 110 & 200 \\
\textbf{Data statistics} & Questions & 228 & 182 & 410 \\
 & Subcategories & 32 & 70 & 94 \\
\midrule
 & Easy & 82 & 90 & 172 \\
\textbf{Difficulty} & Medium & 77 & 57 & 134 \\
 & Hard & 69 & 35 & 104 \\
\bottomrule
\end{tabularx}
\end{table}

Dataset statistics are summarized in Table \ref{tab:composition}. Scenario comparisons are concentrated in the measurement-grounded subtask, whereas the knowledge-grounded subtask covers more physics subcategories. Answer-form composition also varies across difficulty tiers: point estimation is more prominent in Easy questions, and comparisons in Hard questions (Figure \ref{fig:statistics}b,c). Differences in subtask or difficulty scores can therefore reflect a model's strength in particular answer forms alongside its ability to acquire and apply physical information.

Longer and more variable reference solutions in Hard than in Easy questions are consistent with more intermediate dependencies entering the physical calculation in both subtasks (Figure \ref{fig:statistics}e,f). Overlap across tiers nevertheless shows that reference length alone does not characterize construction difficulty. The challenge includes deciding which physical quantities to obtain and how to use them, reinforcing the need to coordinate physical perception, knowledge retrieval, and reasoning. Appendix~\ref{app:construction} gives detailed length summaries and group sizes.

\section{Systematic Evaluation and Failure Analysis}\label{sec:evaluation}

We evaluate models before task-specific training to examine how information demands and solution failures shape their performance. Section \ref{sec:improvement} tests whether trajectory learning improves performance on videos excluded from training.

\subsection{Evaluation Protocol}\label{sec:protocol}

We evaluate GPT-5.5 \cite{openai2026gpt55}, Gemini-3.6-Flash \cite{google2026gemini36flash}, Qwen3.6-27B \cite{qwen2026qwen36_27b}, Qwen3.5-9B \cite{qwen2026qwen35_9b}, Qwen3.5-4B \cite{qwen2026qwen35_4b}, and InteractiveOmni-4B \cite{sensenova2025interactive}. All evaluated models receive the same instructions and solve each question in a native tool loop with web search and Python execution for at most six turns. Each response contains a solution followed by a final answer. The two outputs are assessed separately using the same criteria across models. The full-benchmark comparison covers all 410 questions before task-specific training; the later improvement experiments reuse this protocol on the fixed test partition. Appendix~\ref{app:evaluation} provides generation settings, prompts, and parsing rules.

\subsubsection{Answer Correctness}\label{answer-correctness}

We compare final answers with their references after normalizing numerical values to compatible units. The answer correctness score (ACC) averages question-level scores in \([0,1]\), allowing partial credit. Point estimation is scored by relative error; interval estimation considers overlap and center calibration, and also accepts point predictions inside the reference interval. Physical judgment evaluates the requested yes/no decision, while scenario comparison evaluates both direction and numerical magnitude. Quantitative justification is assessed in the solution score. Appendix~\ref{app:evaluation} specifies all scoring functions and thresholds.

\subsubsection{Solution Correctness and Aggregate Score}\label{solution-correctness-and-aggregate-score}

To complement final-answer evaluation, we assess intermediate reasoning \cite{lightman2023verify,zheng2025processbench} using an LLM judge \cite{zheng2023judging,chen2024mllmjudge}. DeepSeek-V4-Flash \cite{deepseek2026v4flash} compares the model solution with the reference on the identification of quantities and their sources (\(S_{\mathrm{obs}}\)), applicability of the physical relation (\(S_{\mathrm{formula}}\)), validity of numerical transformations (\(S_{\mathrm{derive}}\)), and consistency of the conclusion with the derivation, units, and requested answer (\(S_{\mathrm{conclude}}\)). The solution correctness score (SCS) combines these ratings as

\[
\begin{aligned}
\mathrm{SCS}={}&0.25S_{\mathrm{obs}}+0.25S_{\mathrm{formula}}\\
&+0.30S_{\mathrm{derive}}+0.20S_{\mathrm{conclude}}.
\end{aligned}
\]

Appendix~\ref{app:evaluation} gives the judge prompt and rating rubric.

The overall score (OGS) provides a secondary aggregate:

\[
\mathrm{OGS}=0.70\,\mathrm{ACC}+0.30\,\mathrm{SCS}.
\]

The full-benchmark table reports ACC, three selected solution-score components, and OGS for each subtask. Overall pools all questions and reports ACC, SCS, and OGS. SCS retains all four components, including conclusion consistency with weight 0.20. These component ratings assess the returned solution; they are not separate scores for the three physical capabilities. Higher scores are better; bold identifies column maxima in the comparison tables. Appendices~\ref{app:evaluation} and~\ref{app:results} give aggregation details and subtask-level SCS.

\subsection{Full-Benchmark Comparison}\label{sec:full-results}

All six models struggle to obtain correct physical answers (Table \ref{tab:full-results}). Even Gemini-3.6-Flash, the strongest model by answer correctness, reaches a mean score of only 0.5038. The low answer scores across the evaluated models reveal a persistent gap in coordinating physical perception, physical knowledge retrieval, and physical reasoning. Targeted training must therefore strengthen how these capabilities work together in physical problem solving.

\begin{table*}[pos=!t]
\caption{Full-benchmark evaluation results.}
\label{tab:full-results}
\centering\fontsize{8}{9.5}\selectfont
\setlength{\tabcolsep}{2.7pt}
\renewcommand{\arraystretch}{1.20}
\begin{tabularx}{\textwidth}{@{}>{\raggedright\arraybackslash}X*{5}{r}@{\hspace{7pt}}*{5}{r}@{\hspace{7pt}}*{3}{r}@{}}
\toprule
 & \multicolumn{5}{c}{\textbf{MPR}} & \multicolumn{5}{c}{\textbf{KPR}} & \multicolumn{3}{c}{\textbf{Overall}} \\
\cmidrule(lr){2-6}\cmidrule(lr){7-11}\cmidrule(lr){12-14}
\textbf{Model} & \textbf{ACC} & \(\boldsymbol{S}_{\mathrm{obs}}\) & \(\boldsymbol{S}_{\mathrm{formula}}\) & \(\boldsymbol{S}_{\mathrm{derive}}\) & \textbf{OGS} & \textbf{ACC} & \(\boldsymbol{S}_{\mathrm{obs}}\) & \(\boldsymbol{S}_{\mathrm{formula}}\) & \(\boldsymbol{S}_{\mathrm{derive}}\) & \textbf{OGS} & \textbf{ACC} & \textbf{SCS} & \textbf{OGS} \\
\midrule
\rowcolor[HTML]{FAE6D7}[0pt][0pt]
\multicolumn{14}{@{}c@{}}{\textit{\textbf{Closed-source models}}} \\
\modelicon{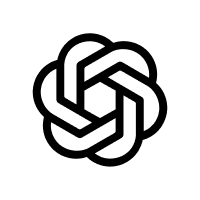}GPT-5.5 & \textbf{0.4316} & 0.4200 & \textbf{0.9287} & 0.6162 & 0.4902 & 0.4597 & 0.5398 & 0.9135 & 0.7143 & 0.5341 & 0.4441 & 0.6627 & 0.5097 \\
\modelicon{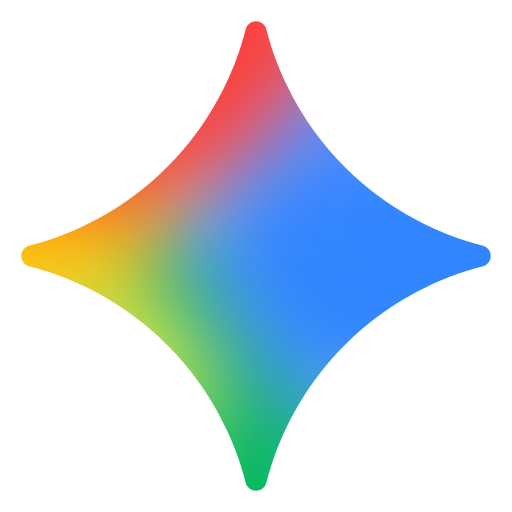}Gemini-3.6-Flash & 0.4267 & \textbf{0.4868} & 0.9002 & \textbf{0.6327} & \textbf{0.4904} & \textbf{0.6004} & \textbf{0.5934} & \textbf{0.9258} & \textbf{0.7486} & \textbf{0.6439} & \textbf{0.5038} & \textbf{0.6862} & \textbf{0.5585} \\
\addlinespace[2pt]
\rowcolor[HTML]{B4DDF4}[0pt][0pt]
\multicolumn{14}{@{}c@{}}{\textit{\textbf{Open-weight models}}} \\
\modelicon{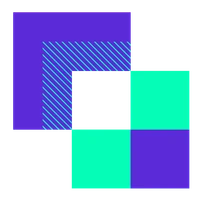}InteractiveOmni-4B & 0.3650 & 0.3531 & 0.8169 & 0.4956 & 0.4097 & 0.2770 & 0.4272 & 0.7115 & 0.5165 & 0.3496 & 0.3259 & 0.5163 & 0.3830 \\
\modelicon{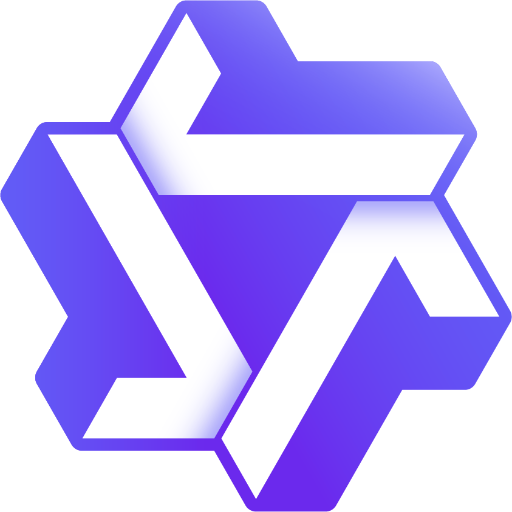}Qwen3.5-4B & 0.3739 & 0.3849 & 0.8246 & 0.5055 & 0.4214 & 0.4151 & 0.5852 & 0.8846 & 0.6648 & 0.4925 & 0.3922 & 0.5947 & 0.4530 \\
\modelicon{qwen}Qwen3.5-9B & 0.3924 & 0.4035 & 0.8750 & 0.5450 & 0.4459 & 0.4536 & 0.5893 & 0.8970 & 0.6937 & 0.5254 & 0.4196 & 0.6249 & 0.4812 \\
\modelicon{qwen}Qwen3.6-27B & 0.4144 & 0.4353 & 0.8904 & 0.5570 & 0.4657 & 0.4672 & 0.5852 & 0.9203 & 0.7005 & 0.5373 & 0.4378 & 0.6366 & 0.4975 \\
\bottomrule
\end{tabularx}
\end{table*}

Answer correctness and solution correctness identify different model strengths within the subtasks. Gemini-3.6-Flash leads in overall ACC, SCS, and OGS, whereas GPT-5.5 achieves the highest ACC and \(S_{\mathrm{formula}}\) on the measurement-grounded subtask. Their MPR OGS remains nearly equal despite this difference in their answer and solution rankings. Evaluating both outputs therefore captures aspects of physical problem solving that a single ranking would obscure.

The measurement-grounded subtask is more demanding in terms of answer correctness for five of the six models, and its ranking differs from that of the knowledge-grounded subtask. GPT-5.5 leads on the former, whereas Gemini-3.6-Flash leads on the latter, where its ACC is 17.4 points higher than on the former. InteractiveOmni-4B is the only model with lower ACC on the knowledge-grounded subtask. Observation ratings are also lower on the measurement-grounded subtask for every model. This change in ordering makes subtask performance important for identifying which models handle the respective information demands more effectively. The subtasks also differ in video sources, physical phenomena, and answer-form composition, so the contrast characterizes their task conditions as a whole.

Among the Qwen configurations, the ordering follows model size. Qwen3.6-27B exceeds Qwen3.5-9B, and Qwen3.5-9B exceeds Qwen3.5-4B, in all three overall metrics, with OGS of 0.4975, 0.4812, and 0.4530. Larger model scale is generally associated with stronger coordination of physical perception, physical knowledge retrieval, and physical reasoning.

\subsection{SCS Scoring Consistency}\label{sec:scs-consistency}

\begin{figure*}[pos=!t]
\centering
\includegraphics[width=\textwidth]{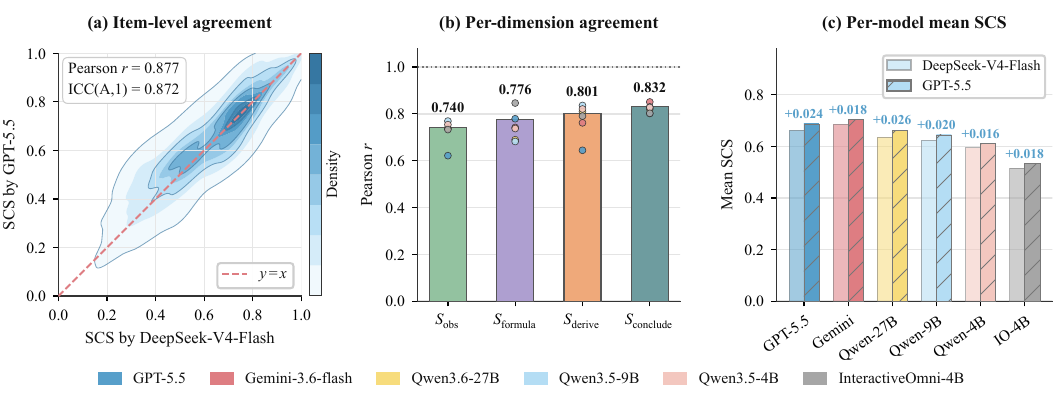}
\caption{SCS agreement between DeepSeek-V4-Flash and GPT-5.5 on 2,460 question--model pairs: (a) joint density of response-level SCS, with V4-Flash on the horizontal axis, GPT-5.5 on the vertical axis, and the dashed line \(y=x\); (b) Pearson correlation of the four components over all pairs (bars) and per evaluated model (dots); (c) mean SCS per model, with the GPT-5.5 minus V4-Flash difference above each bar.}
\label{fig:scs-overview}
\end{figure*}

SCS relies on component ratings from an LLM judge. To examine whether these ratings depend on the judge, GPT-5.5, from a different model family, independently rates all zero-shot responses of the six models on the 410 questions with the same rating prompt, giving 2,460 question--model pairs. Figure \ref{fig:scs-overview} compares its ratings with those of DeepSeek-V4-Flash.

\textbf{Response-level agreement.} The two judges agree closely on individual responses (Figure \ref{fig:scs-overview}a). Their SCS ratings have a Pearson correlation of 0.877 (95\% CI [0.867, 0.886]) and an ICC(A,1) of 0.872, with correlations of 0.881 and 0.856 on the measurement-grounded and knowledge-grounded subtasks. In total, 71.4\% of responses differ by at most 0.10, and only 0.8\% differ by more than 0.25. GPT-5.5 is slightly more lenient, with a mean difference of \(+0.020\); it rates 47\% of responses higher and 30\% lower. Component correlations range from 0.740 for \(S_{\mathrm{obs}}\) to 0.832 for \(S_{\mathrm{conclude}}\) (Figure \ref{fig:scs-overview}b). Identical component ratings occur in 55.7\%--75.3\% of pairs, and fewer than 3\% differ by more than one rating level.

\textbf{Model-level agreement.} The agreement holds for every evaluated model, with SCS correlations of 0.811--0.880. Responses of GPT-5.5 show the lowest correlation, consistent with their narrower score spread (standard deviation 0.141, compared with 0.170--0.203 for the other models). As a judge, GPT-5.5 shows no preference for its own responses: it rates them 0.024 higher than V4-Flash does, within the \(+0.016\) to \(+0.026\) range of the other models (Figure \ref{fig:scs-overview}c). The two judges produce the same model ranking (Spearman \(\rho=1.000\)). Overall, these results indicate that SCS is robust to the choice of LLM judge across model families: response-level and component-level agreement is high, model-level rankings are identical, and the mild leniency of GPT-5.5 does not alter comparative conclusions. This supports the use of SCS as a reliable evaluation signal.

\subsection{Failure Analysis}\label{sec:failure}

\begin{figure*}[pos=!t]
\centering
\includegraphics[width=\textwidth]{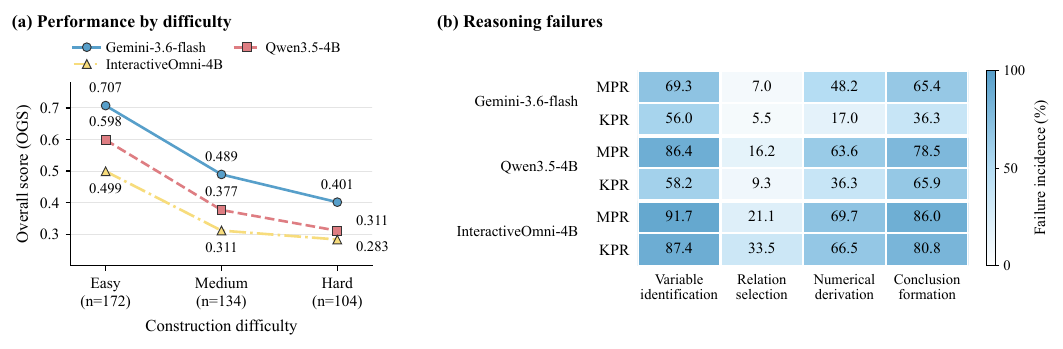}
\caption{Failure analysis: (a) performance by construction difficulty, measured by OGS; (b) component failure rates (ratings of at most 0.5). Rates use all questions in each subtask; categories can overlap.}
\label{fig:failure}
\end{figure*}

We examine performance across construction difficulty and failures within the solutions of Gemini-3.6-Flash, Qwen3.5-4B, and InteractiveOmni-4B. Figure \ref{fig:failure} presents these complementary views on the full benchmark.

\textbf{Sensitivity to difficulty.} All three models score progressively lower from Easy to Medium to Hard questions (Figure \ref{fig:failure}a). Their OGS decreases from 0.707 to 0.401 for Gemini-3.6-Flash, from 0.598 to 0.311 for Qwen3.5-4B, and from 0.499 to 0.283 for InteractiveOmni-4B. The same monotonic decrease holds for all six evaluated models. Performance therefore declines as more information must be acquired before reasoning can be completed, consistent with the construction difficulty assigned to each question.

\textbf{Where solutions fail.} We identify a component failure when its SCS rating is at most 0.5. Incidence is calculated over all questions in each subtask (228 and 182), including those with correct final answers, and categories can overlap. Appendices~\ref{app:evaluation} and~\ref{app:results} provide the definition and underlying counts.

Variable identification is a recurring weakness across all model--subtask pairs, with failure incidence of 56.0\%--91.7\% (Figure \ref{fig:failure}b). This common difficulty concerns which quantities enter a solution and whether each is given, observed in the video, or retrieved. It makes the acquisition of inputs an important training target alongside the selection of a physical relation.

Conclusion formation reveals a more model-dependent pattern. It exceeds derivation failures for every model--subtask pair and is the most frequent failure for Qwen3.5-4B on the knowledge-grounded subtask. Failures also occur in numerical derivation and, least often, in relation selection (5.5\%--33.5\%). The combined profile points to weaknesses in selecting the inputs to a calculation and carrying them through to a supported conclusion.

\subsection{From Diagnosis to Training Requirements}\label{sec:diagnosis-to-training}

These findings motivate learning the dependencies between information acquisition and physical reasoning. We use supervised fine-tuning (SFT) on complete tool-integrated trajectories to supervise how observations, specifications, and calculations enter a solution. Preference optimization then compares complete accepted and rejected candidates, complementing imitation of verified trajectories with an outcome-based signal. Section \ref{sec:triphys} describes this training procedure, and Section \ref{sec:stages} evaluates its successive stages.

\section{TriPhys}\label{sec:triphys}

\begin{figure*}[pos=!t]
\centering
\includegraphics[width=\textwidth]{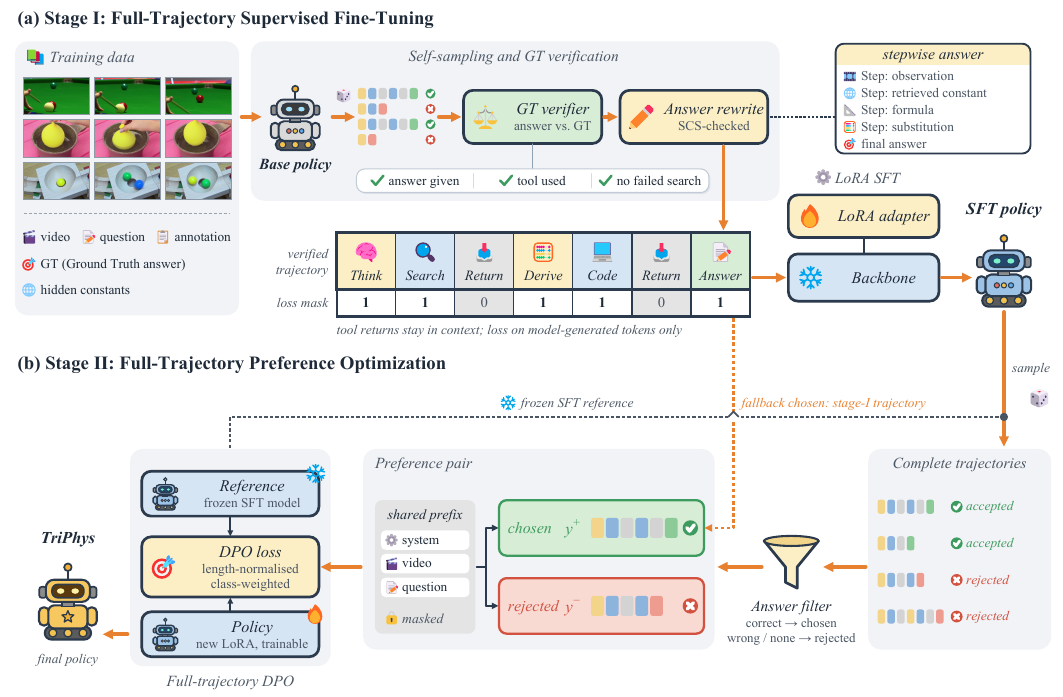}
\caption{TriPhys training: (a) full-trajectory SFT on verified trajectories, with the answer block rewritten into stepwise form; (b) full-trajectory DPO on chosen and rejected trajectories sampled from the SFT policy. In the loss mask, 1 marks supervised model-generated content and 0 marks gray tool returns kept as context; the shared prefix of each pair is masked. The orange dashed line marks the fallback chosen trajectory from Stage I, and the gray dashed line marks the frozen SFT reference; snowflakes denote frozen weights and flames denote trainable LoRA parameters.}
\label{fig:method}
\end{figure*}

TriPhys is a tool-integrated framework for improving physical problem solving in video-capable language models. Given a video and question, the agent identifies the target and required observations, obtains missing physical knowledge when needed, and combines the available quantities with an applicable relation. Numerical tools support the calculation, and the agent uses their returns to form an answer with the requested units and format. We train this behavior through supervised fine-tuning on full trajectories followed by preference optimization. The first stage supervises the full problem-solving sequence of self-sampled solutions that pass answer verification; the second learns from the supervised model's successful and unsuccessful candidate solutions. Figure \ref{fig:method} shows the training procedure.

\subsection{Training Trajectories and Verification}\label{sec:trajectories}

Training questions and their reference answers from the training partition define the supervision. Each solution is serialized as a tool-integrated reasoning trajectory,

\[
\tau_i=\bigl((d_i^{(t)},a_i^{(t)},o_i^{(t)})\bigr)_{t=1}^{T_i},
\]

where \(d_i^{(t)}\) is a model-generated decision or solution segment, \(a_i^{(t)}\) is a tool call or answer action, and \(o_i^{(t)}\) is the corresponding tool observation when present. The serialized sequence records retrieval queries, calculation arguments, returned values, subsequent derivation, and the requested answer. It therefore exposes how an intermediate physical quantity is obtained and used. Tool calls follow the native tool format of the model, which is shared by training, sampling, and evaluation.

A typical sequence identifies a target quantity, selects an applicable relation, acquires a missing property, performs a calculation, and states the result with its unit. The required operations depend on the question: a given motion measurement need not be re-estimated, while a hidden observation or external specification must be obtained before it can support the calculation. This representation preserves the dependencies between the operations needed to solve each question.

The trajectories are sampled first and verified afterward. The backbone solves each training question with tools, without access to the reference answer, and generates \(K\) complete trajectories. A trajectory is retained only when its final answer reaches an answer score of at least \(\delta^{+}\), its last turn contains an answer, it calls at least one tool, and no retrieval fails. For each question, we keep the few retained trajectories with the highest SCS and balance the two subtasks. Because these trajectories come from the model's own distribution, SFT mainly reinforces solution paths that the model can already complete occasionally.

Self-sampled trajectories often end with a bare value, leaving the use of observed and retrieved quantities unstated. A text-only language model therefore rewrites only the final answer block into a stepwise chain, while the reasoning, tool calls, tool observations, and final value remain as generated. The reference solution indicates which quantities and formulas to name, but every value in the chain comes from the trajectory. A rewrite is accepted when its SCS is at least \(\max(s_{\min},\,\mathrm{SCS}_{0})\), where \(\mathrm{SCS}_{0}\) is the score of the original trajectory; otherwise, the original trajectory is retained if \(\mathrm{SCS}_{0}\ge s_{\min}\) and discarded if not.

We generate no separate wording variants. Multiple verified trajectories for the same question vary the query wording, tool order, and reasoning text, while the retrieved value and final answer remain unchanged. This diversity therefore changes expression within the same problem, without adding new physical scenarios.

Two further sources pass the same verification and rewriting. A stronger model solves questions that the backbone never answers correctly, again without access to the reference answer. An intermediate SFT model continues stalled samples, which call tools repeatedly without answering, with a hint to stop retrieval and give an estimate; only correct continuations whose reasoning does not mention the hint are retained, and the hint is removed from the stored trajectory. Appendix~\ref{app:training} details the thresholds and the distinction between solver-generated content and tool observations.

\subsection{Full-Trajectory Supervised Fine-Tuning}\label{sec:sft}

Full-trajectory supervised fine-tuning teaches the agent to solve a physical problem through a complete sequence of decisions, tool calls, derivations, and a final answer \cite{gou2024tora,qin2024toolllm}. We supervise the assistant-generated tokens throughout this sequence. Let \(x_i=(V_i,q_i)\) be the input and let \(y_{i,1:L_i}\) be the serialized target. A mask \(m_{ij}\) equals one for assistant reasoning, tool-call, and rewritten answer tokens, and zero for the system prompt, user input, padding, and external tool-response tokens. The objective is

\[
\mathcal L_{\mathrm{SFT}}=-\frac{1}{N}\sum_{i=1}^{N}
\frac{\sum_{j=1}^{L_i}m_{ij}\log p_\theta(y_{ij}\mid x_i,y_{i,<j})}
{\sum_{j=1}^{L_i}m_{ij}}.
\]

Tool responses remain visible in the conditioning context but are excluded from the prediction loss. This separates the model's responsibility for choosing and using an operation from the external executor's responsibility for returning its result. User tokens are likewise excluded from the prediction target, while remaining available to attention. The loss mask and the attention mask therefore serve different purposes.

The full trajectory provides targets for the dependencies highlighted by the diagnosis: variable references and tool calls identify evidence sources, derivations connect physical relations to numerical transformations, and rewritten answer blocks state the observations, formulas, substitutions, and conclusion in the requested output form.

\subsection{Full-Trajectory Preference Optimization}\label{sec:dpo}

The second stage learns from successful and unsuccessful trajectories sampled from the SFT policy. A final-answer acceptance rule partitions the candidates for each training input. This binary training filter is distinct from the continuous answer correctness score used for evaluation; its thresholds and sampling settings are specified in Section \ref{sec:split}.

For each training input, the SFT policy generates complete trajectories with the same tool loop as in evaluation. Only trajectories whose retrieval returns the reference value of the question enter the candidate pool; trajectories that calculate or answer before any retrieval are discarded. Each of at most a few such prefixes is continued from its first turn that uses the retrieved value, giving \(K\) complete candidates per prefix.

Let \(\mathcal C_i\) contain accepted candidates and \(\mathcal R_i\) contain rejected candidates for input \(x_i\). A candidate is accepted when it gives an answer with ACC of at least \(\delta^{+}\), uses the reference value in its code or answer, and has SCS of at least \(s_{\min}\). It is rejected when it gives no answer or its ACC is below \(\delta^{-}\), with \(\delta^{-}<\delta^{+}\). A preference pair is retained when \(\mathcal R_i\) is nonempty and a chosen solution exists. The chosen and rejected solutions are

\[
\tau_i^+=
\begin{cases}
\underset{\tau\in\mathcal C_i}{\arg\max}\;\mathrm{SCS}(\tau), & \mathcal C_i\neq\varnothing,\\[2pt]
\underset{\tau\in\mathcal S_i}{\arg\max}\;\mathrm{SCS}(\tau), & \mathcal C_i=\varnothing,
\end{cases}
\qquad
\tau_i^-\in\mathcal R_i,
\]

where \(\mathcal S_i\) contains the verified Stage-I trajectories of the question, used as fallback chosen solutions. Each question contributes at most \(M\) pairs, and rejected trajectories alternate between wrong and missing answers. Correctness and solution quality determine this preference rule; trajectory length does not. The comparison uses both complete solutions, which may follow different reasoning paths.

The trainable policy starts from the SFT checkpoint, while a frozen copy serves as the reference policy \(\pi_{\mathrm{ref}}\). With the length-normalized log ratio \(r_\theta(x,\tau)=[\log\pi_\theta(\tau\mid x)-\log\pi_{\mathrm{ref}}(\tau\mid x)]/n(\tau)\), where \(n(\tau)\) is the number of scored assistant tokens, DPO \cite{rafailov2023dpo} minimizes

\[
\mathcal L_{\mathrm{DPO}}=-\mathop{\mathbb E}\limits_{(x,\tau^+,\tau^-)}
w_{c(x)}\log\sigma\bigl(\beta\bar n\,[r_\theta(x,\tau^+)-r_\theta(x,\tau^-)]\bigr).
\]

Here, \(\bar n\) is the mean number of assistant tokens over all preference pairs, keeping the effective temperature \(\beta\bar n\) on the scale of unnormalized DPO. The weight \(w_c=|\mathcal P|/(2|\mathcal P_c|)\) balances the two subtasks, where \(\mathcal P\) contains all preference pairs and \(\mathcal P_c\) those of subtask \(c\). The likelihood is taken over model-generated tokens, conditioned on the preceding input and tool observations; the system prompt, video, and question shared by both trajectories are masked. The reference-relative comparison increases the preference for the selected successful trajectory. It supplies an outcome-based learning signal over complete solutions, whereas SFT imitates the verified target sequence. A correct final answer used for preference selection does not independently verify every intermediate claim; requiring the reference value and a minimum SCS for chosen solutions partly limits the reinforcement of correct answers reached through flawed derivations.

\subsection{Inference}\label{sec:inference}

At inference time, one learned policy receives the video and question, generates solution steps, requests retrieval or computation when needed, and incorporates the returned observations into subsequent steps. Tool executors provide information or numerical outputs; there is no separate planning model. The policy terminates with a final answer in the requested form or at the generation budget. This preserves the same distinction between generated decisions and returned evidence used during training.

\section{Evaluation of TriPhys Models}\label{sec:improvement}

We evaluate TriPhys-4B, TriPhys-9B, and TriPhys-27B, the models obtained by applying the TriPhys framework to three Qwen backbones. We examine their performance on videos excluded from training and the contributions of successive training stages, using the scoring protocol in Section \ref{sec:protocol}.

\subsection{Experimental Setup}\label{sec:split}

\textbf{Data partition.} We split each subtask independently at the video level, keeping all questions from each recording together. The partition contains 300 training questions and 110 test questions (Table \ref{tab:split}). We pool the two training partitions and sample and construct trajectories only on these questions. Appendix~\ref{app:partition} gives the stratification procedure and detailed distributions.

\begin{table}[pos=!t]
\caption{Video-level training and test partitions. Q denotes questions.}
\label{tab:split}
\centering\footnotesize
\setlength{\tabcolsep}{3.5pt}
\renewcommand{\arraystretch}{1.15}
\begin{tabularx}{\columnwidth}{Xrrrr}
\toprule
\textbf{Subtask} & \textbf{Train videos} & \textbf{Train Q} & \textbf{Test videos} & \textbf{Test Q} \\
\midrule
MPR & 66 & 167 & 24 & 61 \\
KPR & 81 & 133 & 29 & 49 \\
Total & 147 & 300 & 53 & 110 \\
\bottomrule
\end{tabularx}
\end{table}

\textbf{Models and baselines.} TriPhys-4B, TriPhys-9B, and TriPhys-27B are initialized from Qwen3.5-4B, Qwen3.5-9B, and Qwen3.6-27B, respectively, and use the same two-stage training procedure. Each is compared with its pretrained counterpart and two closed-source reference models on the same test questions. These questions were included in the full-benchmark diagnosis (Section \ref{sec:evaluation}) but excluded from training.

\textbf{Training configuration.} Base model samples four trajectories per training question, and verification, answer-block rewriting, and the two supplementary sources yield Stage-I trajectories. We use low-rank adaptation (LoRA) \cite{hu2021lora} with rank 32, scaling parameter 64, and dropout 0.05 on the query, key, value, and output attention projections and the gate, up, and down projections, keeping the pretrained weights frozen. SFT runs for three epochs with a learning rate of \(1\times10^{-4}\) and an effective batch size of 16. After SFT, we merge the learned adapter into the base model and train a new adapter for DPO, while keeping the SFT reference fixed. DPO runs for two epochs with a learning rate of \(5\times10^{-6}\), \(\beta=0.1\), and an effective batch size of 16 pairs.

\textbf{Preference construction.} We sample candidates from the SFT checkpoint of TriPhys at temperature 0.7 and top-p 0.9, continuing each of at most two prefixes per training input into four complete candidates. The thresholds are \(\delta^{+}=0.9\), \(\delta^{-}=0.5\), and \(s_{\min}=0.75\), and \(\delta^{+}\) and \(s_{\min}\) also apply to Stage-I verification and rewriting. For point estimates, ACC of at least 0.9 corresponds to a relative error of about 3.3\% or less. We then construct at most two preference pairs per input as described in Section \ref{sec:dpo}. Appendix~\ref{app:training} provides the remaining settings.

\subsection{Results Across Model Sizes}\label{sec:training-results}

All three TriPhys models improve both answer and solution correctness over their pretrained backbones on the fixed test partition (Table \ref{tab:training-results}). Relative answer-score gains range from 16.4\% to 17.7\%. The improvement across model sizes and both evaluation targets supports complete tool-trajectory learning as a practical way to strengthen physical problem solving on videos excluded from training.

\begin{table*}[pos=!t]
\caption{Test-set performance of TriPhys models and baselines. Green marks TriPhys; subscripts show relative gains (\%) over each backbone.}
\label{tab:training-results}
\centering\footnotesize
\setlength{\tabcolsep}{4pt}
\renewcommand{\arraystretch}{1.2}
\begin{tabularx}{\textwidth}{l*{5}{>{\raggedleft\arraybackslash}X}}
\toprule
 & \multicolumn{1}{c}{\textbf{MPR}} & \multicolumn{1}{c}{\textbf{KPR}} & \multicolumn{3}{c}{\textbf{Overall}} \\
\cmidrule(lr){2-2}\cmidrule(lr){3-3}\cmidrule(lr){4-6}
\textbf{Model} & \multicolumn{1}{c}{\textbf{ACC}} & \multicolumn{1}{c}{\textbf{ACC}} & \multicolumn{1}{c}{\textbf{ACC}} & \multicolumn{1}{c}{\textbf{SCS}} & \multicolumn{1}{c}{\textbf{OGS}} \\
\midrule
\rowcolor[HTML]{FAE6D7}
\multicolumn{6}{c}{\textit{\textbf{Closed-source models}}} \\
GPT-5.5 & 0.5171\kern0.35pt{\fontsize{6.5bp}{7bp}\selectfont\hphantom{00.0}} & 0.5043\kern0.35pt{\fontsize{6.5bp}{7bp}\selectfont\hphantom{00.0}} & 0.5114\kern0.35pt{\fontsize{6.5bp}{7bp}\selectfont\hphantom{00.0}} & 0.6907\kern0.35pt{\fontsize{6.5bp}{7bp}\selectfont\hphantom{0.0}} & 0.5652\kern0.35pt{\fontsize{6.5bp}{7bp}\selectfont\hphantom{00.0}} \\
Gemini-3.6-Flash & \textbf{0.5328}\kern0.35pt{\fontsize{6.5bp}{7bp}\selectfont\hphantom{00.0}} & 0.6634\kern0.35pt{\fontsize{6.5bp}{7bp}\selectfont\hphantom{00.0}} & \textbf{0.5910}\kern0.35pt{\fontsize{6.5bp}{7bp}\selectfont\hphantom{00.0}} & 0.7042\kern0.35pt{\fontsize{6.5bp}{7bp}\selectfont\hphantom{0.0}} & \textbf{0.6250}\kern0.35pt{\fontsize{6.5bp}{7bp}\selectfont\hphantom{00.0}} \\
\addlinespace[2pt]
\rowcolor[HTML]{B4DDF4}
\multicolumn{6}{c}{\textit{\textbf{Open-weight models}}} \\
\rowcolor[HTML]{F6F6F6}
\textbf{Qwen3.5-4B} & 0.4412\kern0.35pt{\fontsize{6.5bp}{7bp}\selectfont\hphantom{00.0}} & 0.4932\kern0.35pt{\fontsize{6.5bp}{7bp}\selectfont\hphantom{00.0}} & 0.4644\kern0.35pt{\fontsize{6.5bp}{7bp}\selectfont\hphantom{00.0}} & 0.6251\kern0.35pt{\fontsize{6.5bp}{7bp}\selectfont\hphantom{0.0}} & 0.5126\kern0.35pt{\fontsize{6.5bp}{7bp}\selectfont\hphantom{00.0}} \\
\rowcolor[HTML]{E5F4E8}
\hspace{0.8em}TriPhys-4B & 0.5179\kern0.35pt\raisebox{-2.25bp}{\fontsize{6.5bp}{7bp}\selectfont 17.4} & 0.5729\kern0.35pt\raisebox{-2.25bp}{\fontsize{6.5bp}{7bp}\selectfont 16.2} & 0.5424\kern0.35pt\raisebox{-2.25bp}{\fontsize{6.5bp}{7bp}\selectfont 16.8} & 0.6828\kern0.35pt\raisebox{-2.25bp}{\fontsize{6.5bp}{7bp}\selectfont 9.2} & 0.5845\kern0.35pt\raisebox{-2.25bp}{\fontsize{6.5bp}{7bp}\selectfont 14.0} \\
\midrule
\rowcolor[HTML]{F6F6F6}
\textbf{Qwen3.5-9B} & 0.4117\kern0.35pt{\fontsize{6.5bp}{7bp}\selectfont\hphantom{00.0}} & 0.5405\kern0.35pt{\fontsize{6.5bp}{7bp}\selectfont\hphantom{00.0}} & 0.4690\kern0.35pt{\fontsize{6.5bp}{7bp}\selectfont\hphantom{00.0}} & 0.6618\kern0.35pt{\fontsize{6.5bp}{7bp}\selectfont\hphantom{0.0}} & 0.5268\kern0.35pt{\fontsize{6.5bp}{7bp}\selectfont\hphantom{00.0}} \\
\rowcolor[HTML]{E5F4E8}
\hspace{0.8em}TriPhys-9B & 0.4887\kern0.35pt\raisebox{-2.25bp}{\fontsize{6.5bp}{7bp}\selectfont 18.7} & 0.6169\kern0.35pt\raisebox{-2.25bp}{\fontsize{6.5bp}{7bp}\selectfont 14.1} & 0.5458\kern0.35pt\raisebox{-2.25bp}{\fontsize{6.5bp}{7bp}\selectfont 16.4} & 0.7016\kern0.35pt\raisebox{-2.25bp}{\fontsize{6.5bp}{7bp}\selectfont 6.0} & 0.5925\kern0.35pt\raisebox{-2.25bp}{\fontsize{6.5bp}{7bp}\selectfont 12.5} \\
\midrule
\rowcolor[HTML]{F6F6F6}
\textbf{Qwen3.6-27B} & 0.4471\kern0.35pt{\fontsize{6.5bp}{7bp}\selectfont\hphantom{00.0}} & 0.5525\kern0.35pt{\fontsize{6.5bp}{7bp}\selectfont\hphantom{00.0}} & 0.4941\kern0.35pt{\fontsize{6.5bp}{7bp}\selectfont\hphantom{00.0}} & 0.6784\kern0.35pt{\fontsize{6.5bp}{7bp}\selectfont\hphantom{0.0}} & 0.5494\kern0.35pt{\fontsize{6.5bp}{7bp}\selectfont\hphantom{00.0}} \\
\rowcolor[HTML]{E5F4E8}
\hspace{0.8em}TriPhys-27B & 0.5154\kern0.35pt\raisebox{-2.25bp}{\fontsize{6.5bp}{7bp}\selectfont 15.3} & \textbf{0.6646}\kern0.35pt\raisebox{-2.25bp}{\fontsize{6.5bp}{7bp}\selectfont 20.3} & 0.5818\kern0.35pt\raisebox{-2.25bp}{\fontsize{6.5bp}{7bp}\selectfont 17.7} & \textbf{0.7156}\kern0.35pt\raisebox{-2.25bp}{\fontsize{6.5bp}{7bp}\selectfont 5.5} & 0.6219\kern0.35pt\raisebox{-2.25bp}{\fontsize{6.5bp}{7bp}\selectfont 13.2} \\
\bottomrule
\end{tabularx}
\end{table*}

Both subtasks benefit in all three models. The absolute answer-score gains are similar across subtasks for TriPhys-4B and TriPhys-9B, whereas TriPhys-27B gains more on the knowledge-grounded subtask. For TriPhys-4B and TriPhys-9B, the larger relative gains on the measurement-grounded subtask also reflect its lower starting scores. The overall improvement thus extends across both information settings, while the subtask breakdown distinguishes the size of the gain from the difficulty of the starting point.

Task-specific adaptation also changes the comparison with closed-source reference models. All three TriPhys models exceed GPT-5.5's mean answer score of 51.14\% on the same test questions, and TriPhys-27B attains 58.18\%. TriPhys-27B also leads in solution correctness and is on par with Gemini-3.6-Flash in knowledge-grounded answer correctness, but remains below Gemini-3.6-Flash in overall answer correctness and the overall score. These results establish the benefit of adaptation within this evaluation, while the remaining answer-score deficit shows that the tasks are still challenging.

After training, performance follows model size in all three overall metrics. TriPhys-4B and TriPhys-9B differ by only 0.003 in answer correctness, as their backbones already differ by only 0.005. These configurations span different releases, so the evidence is strongest for the matched before/after comparisons.

\subsection{Training-Stage Ablation}\label{sec:stages}

\textbf{Setup.} We examine the training stages of TriPhys-4B, TriPhys-9B, and TriPhys-27B through three checkpoints each: the pretrained backbone (Base), the checkpoint after SFT on verified trajectories, and the full model after subsequent DPO. All nine checkpoints are evaluated on the same 110 test questions with the same scoring protocol. This design compares the full recipe with its intermediate checkpoints (Table \ref{tab:stages}).

\begin{table}[pos=!t]
\caption{Training-stage ablation of TriPhys models. Gray marks the pretrained backbones and green marks the trained stages. Increment denotes the OGS gain over the preceding stage; bold marks the best value for each model.}
\label{tab:stages}
\centering\footnotesize
\setlength{\tabcolsep}{2pt}
\renewcommand{\arraystretch}{1.15}
\begin{tabularx}{\columnwidth}{Xrrrrrr}
\toprule
 & \multicolumn{1}{c}{\textbf{MPR}} & \multicolumn{1}{c}{\textbf{KPR}} & \multicolumn{3}{c}{\textbf{Overall}} &  \\
\cmidrule(lr){2-2}\cmidrule(lr){3-3}\cmidrule(lr){4-6}
\textbf{Stage} & \textbf{ACC} & \textbf{ACC} & \textbf{ACC} & \textbf{SCS} & \textbf{OGS} & \textbf{Increment} \\
\midrule
\rowcolor[HTML]{B4DDF4}
\multicolumn{7}{c}{\textit{\textbf{TriPhys-4B}} \textit{(Qwen3.5-4B backbone)}} \\
\rowcolor[HTML]{F6F6F6}
\textbf{Base} & 0.4412 & 0.4932 & 0.4644 & 0.6251 & 0.5126 & -- \\
\rowcolor[HTML]{E5F4E8}
\hspace{0.8em}SFT & 0.4740 & 0.5544 & 0.5098 & 0.6818 & 0.5614 & +0.0488 \\
\rowcolor[HTML]{E5F4E8}
\hspace{0.8em}SFT + DPO & \textbf{0.5179} & \textbf{0.5729} & \textbf{0.5424} & \textbf{0.6828} & \textbf{0.5845} & +0.0231 \\
\addlinespace[2pt]
\rowcolor[HTML]{B4DDF4}
\multicolumn{7}{c}{\textit{\textbf{TriPhys-9B}} \textit{(Qwen3.5-9B backbone)}} \\
\rowcolor[HTML]{F6F6F6}
\textbf{Base} & 0.4117 & 0.5405 & 0.4690 & 0.6618 & 0.5268 & -- \\
\rowcolor[HTML]{E5F4E8}
\hspace{0.8em}SFT & 0.4625 & \textbf{0.6233} & 0.5341 & 0.6728 & 0.5757 & +0.0489 \\
\rowcolor[HTML]{E5F4E8}
\hspace{0.8em}SFT + DPO & \textbf{0.4887} & 0.6169 & \textbf{0.5458} & \textbf{0.7016} & \textbf{0.5925} & +0.0168 \\
\addlinespace[2pt]
\rowcolor[HTML]{B4DDF4}
\multicolumn{7}{c}{\textit{\textbf{TriPhys-27B}} \textit{(Qwen3.6-27B backbone)}} \\
\rowcolor[HTML]{F6F6F6}
\textbf{Base} & 0.4471 & 0.5525 & 0.4941 & 0.6784 & 0.5494 & -- \\
\rowcolor[HTML]{E5F4E8}
\hspace{0.8em}SFT & 0.4570 & 0.5410 & 0.4944 & 0.7051 & 0.5576 & +0.0082 \\
\rowcolor[HTML]{E5F4E8}
\hspace{0.8em}SFT + DPO & \textbf{0.5154} & \textbf{0.6646} & \textbf{0.5818} & \textbf{0.7156} & \textbf{0.6219} & +0.0643 \\
\bottomrule
\end{tabularx}
\end{table}

\textbf{Results.} Neither training stage lowers overall answer correctness in any configuration. For TriPhys-4B and TriPhys-9B, SFT produces the larger initial improvement, and subsequent DPO adds 0.033 and 0.012 in answer correctness. For TriPhys-27B, SFT leaves answer correctness unchanged while raising solution correctness, and DPO raises answer correctness by 0.087, with gains in both subtasks. The gains support a recipe that first learns complete verified solutions and then refines preferences among generated alternatives. They quantify the benefit of adding preference learning to the supervised checkpoint; independent effects of answer-block rewriting and loss masking remain untested. Appendix~\ref{app:results} provides the detailed increments and their calculation.

\subsection{Case Study}\label{sec:case-study}

\begin{figure*}[pos=!t]
\centering
\includegraphics[width=0.92\textwidth]{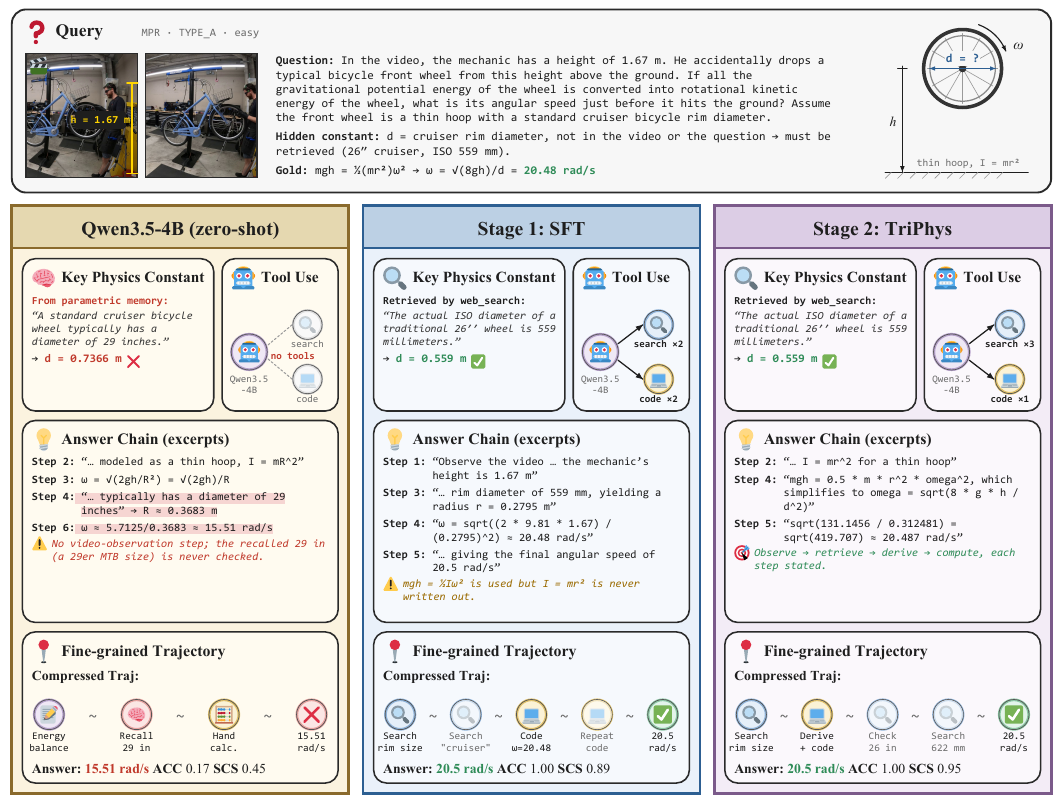}
\caption{Case study of TriPhys-4B on an MPR test question. The columns show Qwen3.5-4B, the SFT model, and the TriPhys model. Each column lists the source of the rim diameter, the tool calls, excerpts of the answer chain, a compressed trajectory, and the answer with ACC and SCS. Red highlights mark errors, and faded nodes mark tool calls whose results are not used.}
\label{fig:case}
\end{figure*}

Figure \ref{fig:case} illustrates how the solution of TriPhys-4B changes across its three stages on an Easy MPR point-estimation question. A bicycle front wheel drops from a mechanic's height of 1.67 m, and the question asks for its angular speed when all gravitational potential energy becomes rotational kinetic energy, treating the wheel as a thin hoop with a standard cruiser rim diameter. The relation \(\omega=\sqrt{8gh}/d\) is simple, but the rim diameter is neither given in the question nor measurable from the video; the reference retrieves the 559 mm ISO diameter of a 26-inch cruiser wheel and obtains 20.48 rad/s. For example, the Base model writes the correct energy balance and thin-hoop moment of inertia, receiving 0.75 for \(S_{\mathrm{formula}}\), but recalls a diameter of 29 inches, a mountain-bike size, and answers 15.51 rad/s. Substituting 0.559 m into its own derivation gives about 20.4 rad/s, so the error comes from this single constant. After SFT, the model retrieves the constant before calculating. Its first query returns the 559 mm diameter, which the solution states together with the query, and the code executor gives 20.48 rad/s; the 507 mm BMX cruiser rim returned by a second query is not adopted. The solution follows observation, modeling, retrieval, and calculation, but it never writes out \(I=mr^2\) and repeats the same calculation in two code runs. After DPO, the retrieval and answer are unchanged, while the derivation becomes complete: the solution states \(I=mr^2\), simplifies the energy balance to the reference form \(\omega=\sqrt{8gh/d^2}\), and shows the substituted values, raising SCS from 0.888 to 0.950.

The DPO output still contains minor errors. Its handwritten product \(8\times9.81\times1.67\) is 131.1456 rather than 131.0616, although the code result and the rounded answer are correct, and it attributes the 559 mm value to a later query. After the calculation, it also issues two further searches, which return a product listing and the 622 mm road-bike rim, and adopts neither value. Overall, SFT replaces a recalled constant with a retrieved and stated one, raising ACC from 0.168 to 1.000, while DPO makes the derivation explicit without changing the answer; the ACC gain of DPO in Table \ref{tab:stages} comes from other questions.

\subsection{Sensitivity Analysis}\label{sec:sensitivity}

\begin{figure}[pos=!t]
\centering
\includegraphics[width=\columnwidth]{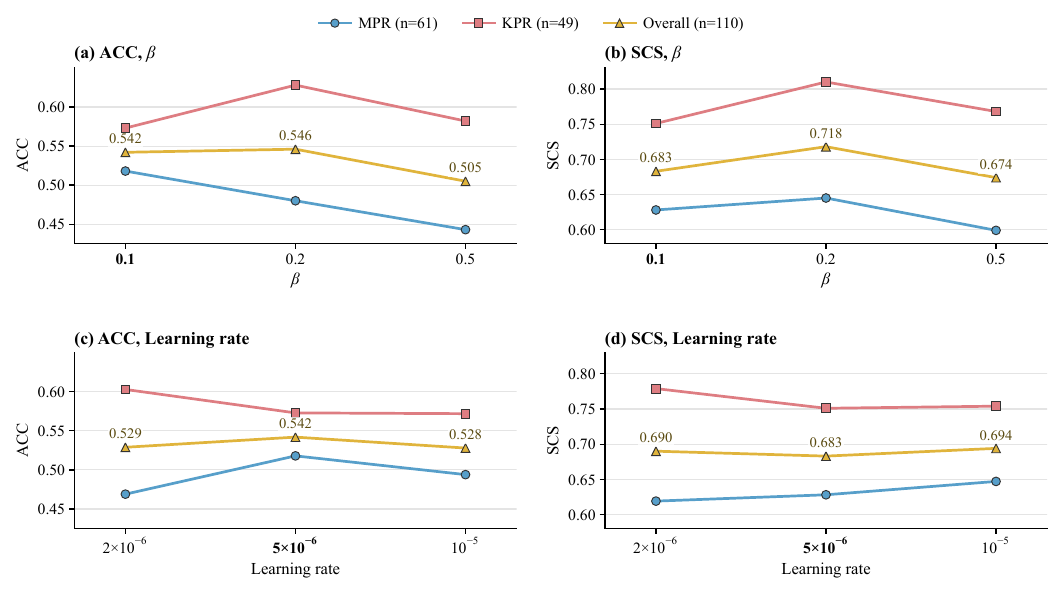}
\caption{Sensitivity of TriPhys-4B to preference-optimization hyperparameters: (a) ACC and (b) SCS for \(\beta\); (c) ACC and (d) SCS for the learning rate. One hyperparameter changes at a time; bold ticks mark the defaults. Blue, pink, and yellow denote MPR (61 questions), KPR (49), and Overall (110), with Overall values annotated.}
\label{fig:sensitivity}
\end{figure}

\textbf{Setup.} We vary one preference-optimization hyperparameter at a time for TriPhys-4B while keeping the SFT checkpoint and the remaining settings fixed. The coefficient \(\beta\) takes values of \(0.1\), \(0.2\), and \(0.5\), and the learning rate takes values of \(2\times10^{-6}\), \(5\times10^{-6}\), and \(1\times10^{-5}\), with defaults of \(\beta=0.1\) and \(5\times10^{-6}\). Each variant is evaluated on the same 110 test questions (Figure~\ref{fig:sensitivity}).

\textbf{Results.} The overall metrics exhibit only modest sensitivity to these hyperparameters within the tested ranges. For \(\beta\) of \(0.1\)--\(0.2\) and learning rates up to \(1\times10^{-5}\), overall ACC remains within \(0.528\)--\(0.546\) and SCS within \(0.683\)--\(0.718\). The overall ACC differences from the default are at most \(0.014\) in this range and are below the standard error of about \(0.048\) for 110 questions. A \(\beta\) of \(0.2\) raises overall ACC by \(0.004\) and SCS by \(0.035\), but this gain comes with a subtask trade-off: MPR ACC decreases by \(0.038\), while KPR ACC increases by \(0.055\). Stronger updates mainly affect the measurement-grounded subtask. A learning rate of \(2\times10^{-6}\) lowers MPR ACC to \(0.469\), \(0.049\) below the default, while raising KPR ACC by \(0.030\); a learning rate of \(1\times10^{-5}\) lowers MPR ACC by \(0.024\) and leaves KPR ACC nearly unchanged (\(-0.001\)). A large \(\beta\) of \(0.5\) lowers MPR ACC by \(0.075\) and overall ACC by \(0.037\), while KPR ACC rises by \(0.009\), giving an overall ACC of \(0.505\), SCS of \(0.674\), and OGS of \(0.5557\). We therefore retain \(\beta=0.1\) and a learning rate of \(5\times10^{-6}\) as the default configuration; within the tested ranges, the main conclusions are not materially sensitive to these hyperparameters.

\section{Discussion and Limitations}\label{sec:discussion}

TriPhysBench makes the acquisition and use of physical information a shared demand of agent evaluation. Low baseline answer scores and improvements after TriPhys training show both the difficulty of these tasks and the value of learning from complete physical solutions. Answer and solution scores provide complementary views, while component-level diagnosis identifies weaknesses that an overall ranking would conceal.

Our evaluation covers six VLMs, and TriPhys training is tested on three Qwen configurations. Broader comparisons across model families, capacities, and tool-use strategies would establish how the observed capability gaps and training gains vary with model design. Repeated runs would help quantify the stability of these gains.

The current scores assess how the three physical capabilities work together. Matched questions could probe their dependencies more directly by varying the observations or specifications supplied for the same physical problem. Analysis of tool trajectories, supported by human validation of intermediate quantities and of model-based ratings, which can carry evaluator biases \cite{wang2024fair}, would help distinguish failures in information acquisition from errors in applying available evidence. Comparing these diagnoses before and after training could reveal which capabilities improve and where errors continue to propagate.

\section{Conclusion}\label{sec:conclusion}

We introduced TriPhysBench to evaluate physical intelligence for embodied agents through the coordination of physical perception, physical knowledge retrieval, and physical reasoning. Its 410 questions from 200 videos span two complementary subtasks with different observation and knowledge requirements. Evaluation of six models reveals low answer scores under these joint demands, and detailed analysis identifies weaknesses in the use of quantities and the formation of supported conclusions. TriPhys integrates information acquisition and computation into physical problem solving and improves all three evaluated Qwen configurations on 110 questions from held-out videos. TriPhys-27B reaches an answer correctness score of 0.5818 and an overall score of 0.6219, with a 17.7\% relative improvement in answer correctness. The cumulative stage comparisons of the three configurations show overall-score gains after supervised training and further gains after preference optimization. Together, these contributions provide an evaluation foundation and a learning pathway for embodied agents to progress from recognizing physical events to understanding and reasoning about the processes that govern them.

\section*{Acknowledgments}
The authors used OpenAI GPT models \cite{openai2026textgeneration} to polish the manuscript's language and GPT Image \cite{openai2026imagegeneration} to generate illustrative icons and schematic elements in Figures~\ref{fig:overview} and~\ref{fig:construction}. The research concepts, experimental design, interpretation of results, and conclusions were developed by the authors. All AI-assisted material was reviewed and revised by the authors, who take full responsibility for the final manuscript. AI use in the research workflow is described in Sections~\ref{sec:construction}--\ref{sec:improvement}.

\bibliographystyle{IEEEtran}
\bibliography{references,LingPhysKnow_preview_additions,TriPhysBench_new_references}

\FloatBarrier\newpage
\section*{Author Biographies}

\Needspace{32mm}
\noindent\begin{minipage}[t]{0.24\linewidth}\vspace{0pt}
\includegraphics[width=18mm,height=23mm,trim=0 101.25bp 0 0,clip,keepaspectratio]{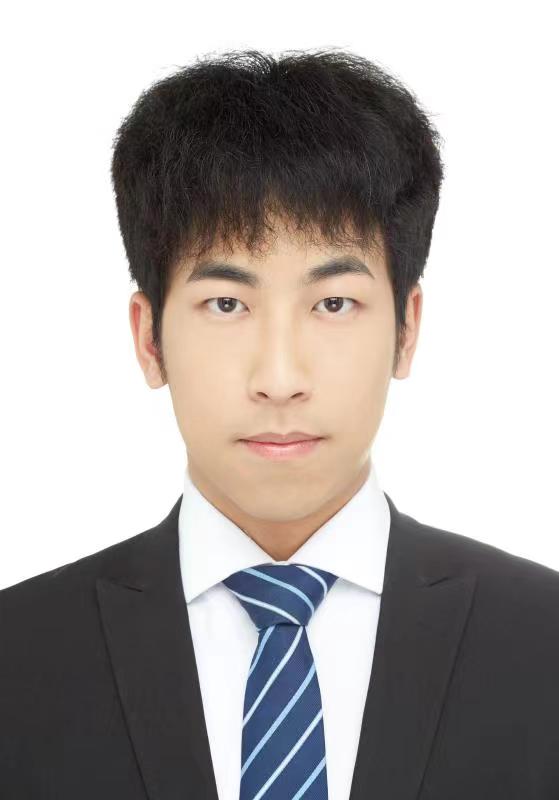}
\end{minipage}\hfill
\begin{minipage}[t]{0.72\linewidth}\vspace{0pt}\fontsize{8}{10.2}\selectfont
\textbf{Zhongxiang Sun}\enspace received the B.Eng. degree in computer science and technology from Beijing Jiaotong University in 2022. He is currently pursuing the Ph.D. degree at the Gaoling School of Artificial Intelligence, Renmin University of China, under the supervision of Prof. Jun Xu. His research interests include trustworthy large language models and agents, mechanistic interpretability, and retrieval-augmented reasoning. His work has appeared in ICLR, SIGIR, and CIKM, including an ICLR 2025 Spotlight paper. He received Best Paper Awards at SIGIR-AP 2024 and CCIR 2026.
\end{minipage}\par\medskip

\Needspace{32mm}
\noindent\begin{minipage}[t]{0.24\linewidth}\vspace{0pt}
\includegraphics[width=18mm,height=23mm,trim=0 101.25bp 0 0,clip,keepaspectratio]{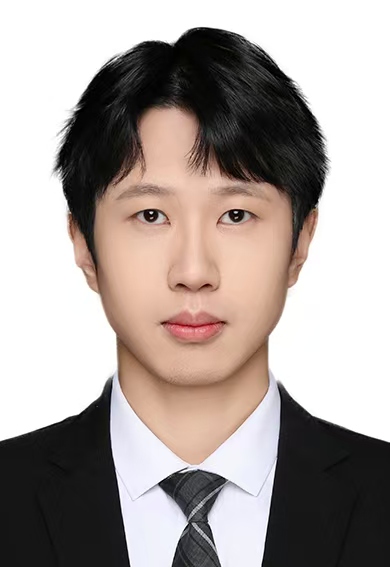}
\end{minipage}\hfill
\begin{minipage}[t]{0.72\linewidth}\vspace{0pt}\fontsize{8}{10.2}\selectfont
\textbf{Chenghao Yang}\enspace received the M.S. degree in data science from The University of Sydney in 2023. He is currently pursuing the Ph.D. degree at the School of Information Engineering, Zhongnan University of Economics and Law. His research interests include data mining, AI for Time Series, and trustworthy large language models and agents.
\end{minipage}\par\medskip

\Needspace{32mm}
\noindent\begin{minipage}[t]{0.24\linewidth}\vspace{0pt}
\includegraphics[width=18mm,height=23mm,trim=539.8bp 0 539.8bp 0,clip,keepaspectratio]{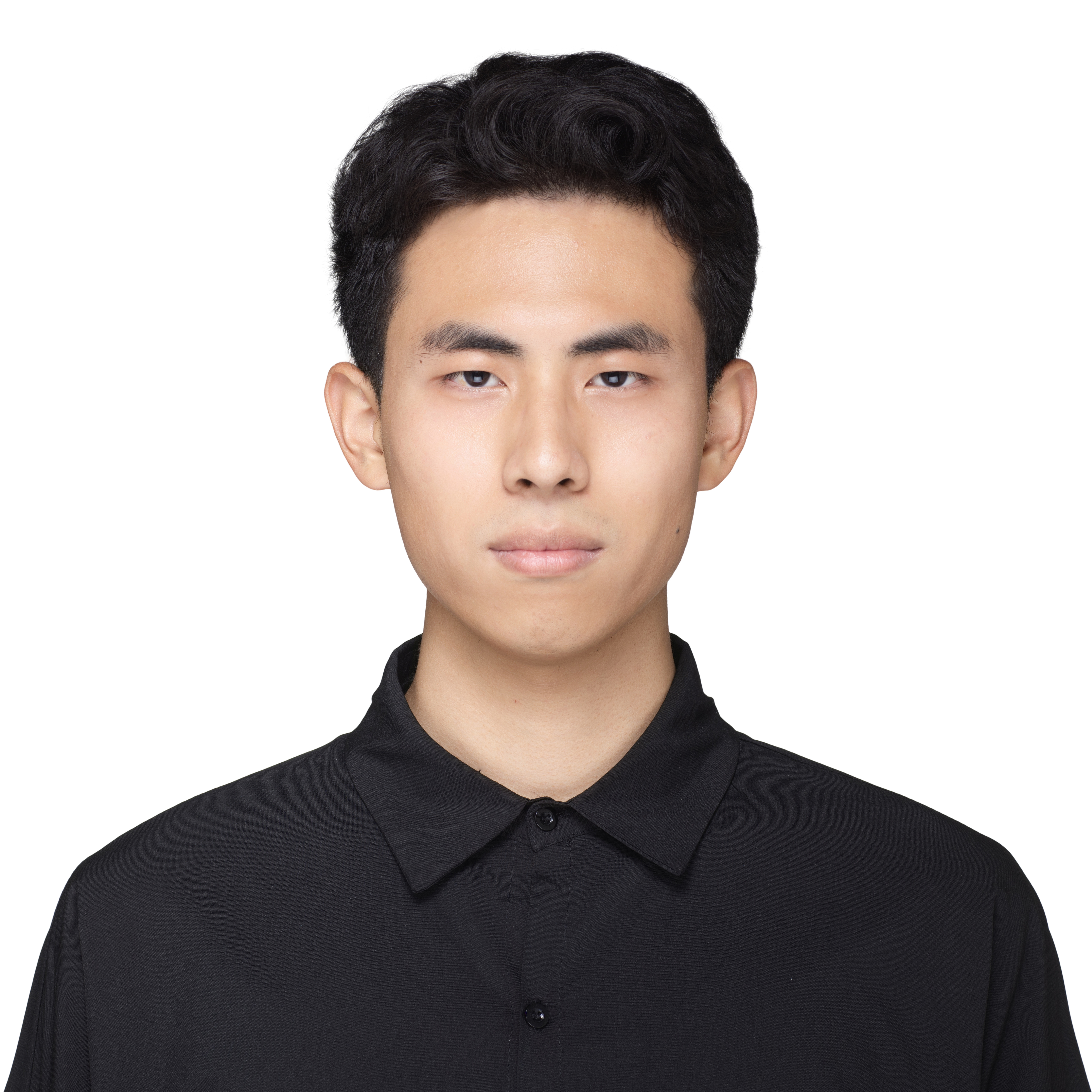}
\end{minipage}\hfill
\begin{minipage}[t]{0.72\linewidth}\vspace{0pt}\fontsize{8}{10.2}\selectfont
\textbf{Haojie Ding}\enspace is a Researcher at Kuaishou Technology. His research interests include large language models, multimodal learning, recommender systems, and AI agents. His recent work covers video understanding, reasoning-based content moderation, and generative recommendation. He has contributed to the Kwai Keye-VL multimodal foundation model and the application of vision-language reasoning to content moderation on short-video platforms.
\end{minipage}\par\medskip

\Needspace{32mm}
\noindent\begin{minipage}[t]{0.24\linewidth}\vspace{0pt}
\includegraphics[width=18mm,height=23mm,trim=0 63bp 0 0,clip,keepaspectratio]{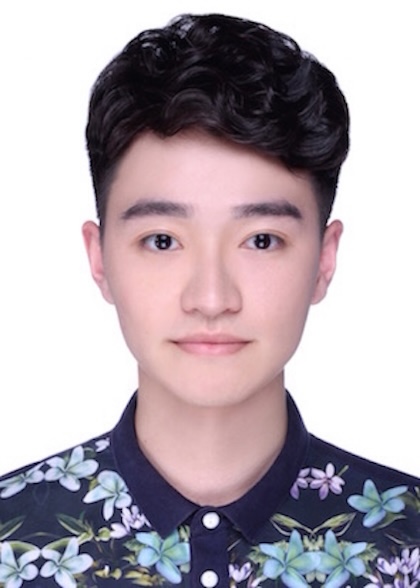}
\end{minipage}\hfill
\begin{minipage}[t]{0.72\linewidth}\vspace{0pt}\fontsize{8}{10.2}\selectfont
\textbf{Boheng Zhang}\enspace is a Senior Algorithm Expert at Kuaishou Technology. His research focuses on multimodal foundation models and image and video generation. He has published more than 10 papers in venues including NeurIPS, EMNLP, ACM MM, CVPR, and Bioinformatics. He also serves as a reviewer for leading conferences and journals in the field.
\end{minipage}\par\medskip

\Needspace{32mm}
\noindent\begin{minipage}[t]{0.24\linewidth}\vspace{0pt}
\includegraphics[width=18mm,height=23mm,trim=141.6bp 0 141.6bp 0,clip,keepaspectratio]{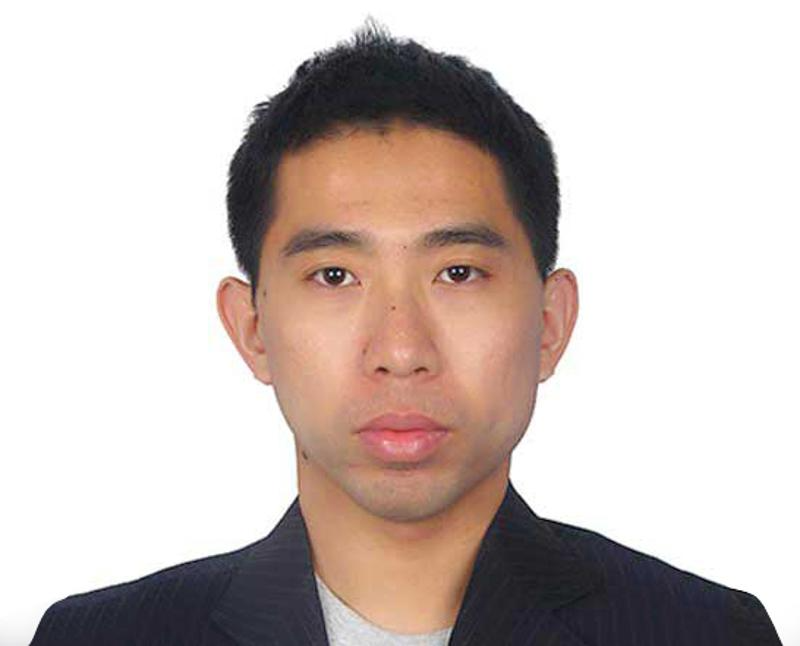}
\end{minipage}\hfill
\begin{minipage}[t]{0.72\linewidth}\vspace{0pt}\fontsize{8}{10.2}\selectfont
\textbf{Fan Yang}\enspace is a Senior Algorithm Expert at Kuaishou Technology. His research focuses on large language models, multimodal intelligence, and agentic AI systems, with an emphasis on foundation models and their applications. His research has been published in ICLR, NeurIPS, ICML, KDD, AAAI, and EMNLP. His work spans the development and deployment of large-scale AI systems in industrial applications.
\end{minipage}\par\medskip

\Needspace{32mm}
\noindent\begin{minipage}[t]{0.24\linewidth}\vspace{0pt}
\includegraphics[width=18mm,height=23mm,trim=0 20bp 0 0,clip,keepaspectratio]{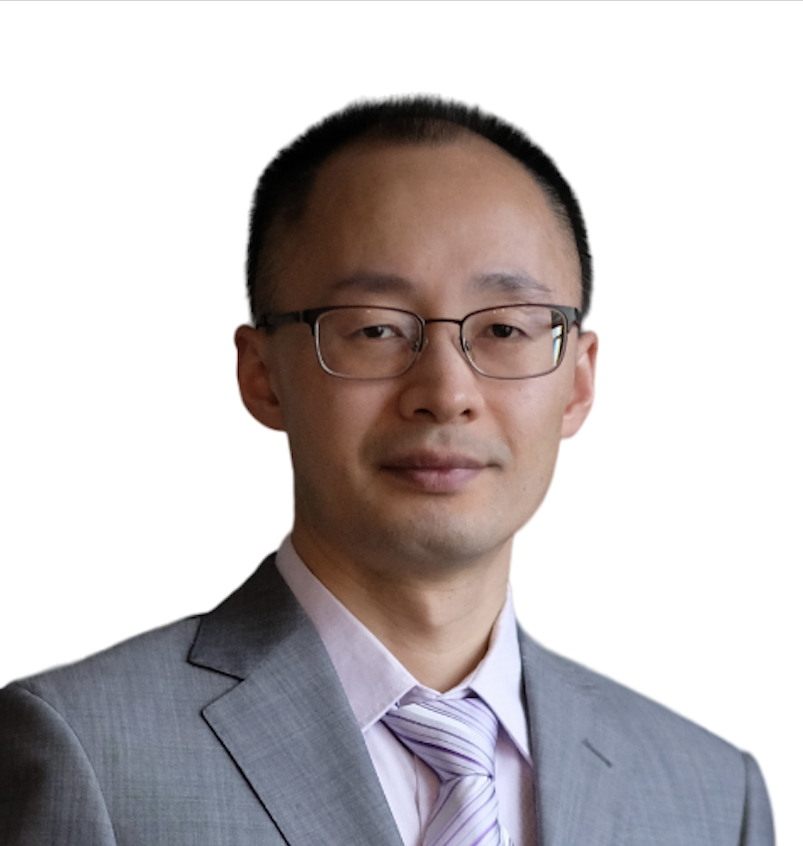}
\end{minipage}\hfill
\begin{minipage}[t]{0.72\linewidth}\vspace{0pt}\fontsize{8}{10.2}\selectfont
\textbf{Qingsong Wen}\enspace (Senior Member, IEEE) is currently the Chief Scientist at Squirrel Ai Learning (a top EdTech unicorn with 3,000+ learning centers across 1,500+ cities, serving tens of millions of students globally), and PhD Supervisor at University of Oxford. Before that, he worked at Alibaba, Qualcomm, Marvell, etc., and received his M.S. and Ph.D. degrees in Electrical and Computer Engineering from Georgia Institute of Technology, USA. His research interests include machine learning, data mining, and signal processing, especially LLM \& AI Agent, AI for Education, and AI for Time Series.
\end{minipage}\par\medskip

\Needspace{32mm}
\noindent\begin{minipage}[t]{0.24\linewidth}\vspace{0pt}
\includegraphics[width=18mm,height=23mm,trim=80bp 40bp 80bp 20bp,clip,keepaspectratio]{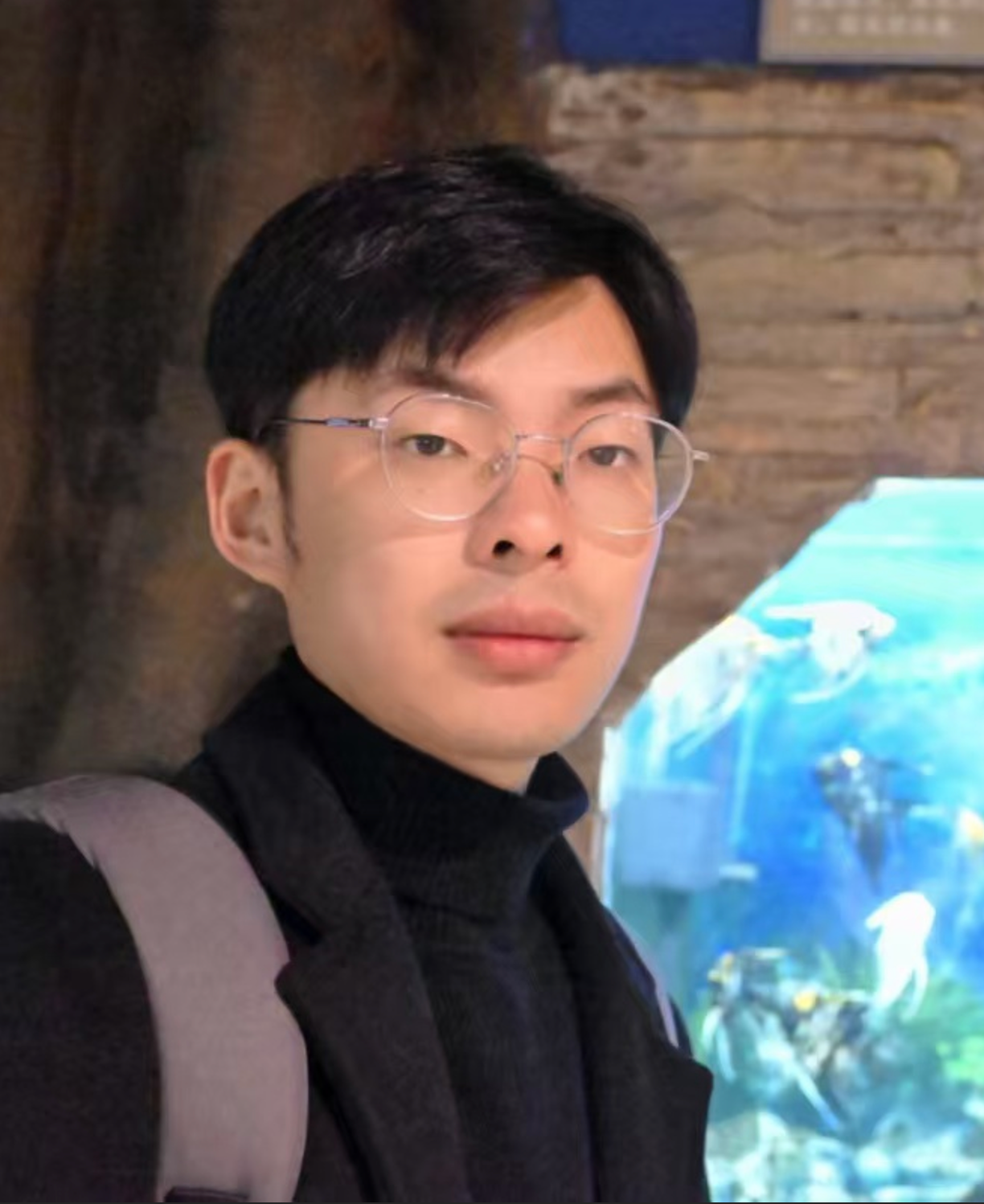}
\end{minipage}\hfill
\begin{minipage}[t]{0.72\linewidth}\vspace{0pt}\fontsize{8}{10.2}\selectfont
\textbf{Kun Wang}\enspace is a Professor at the School of Artificial Intelligence and Automation, Huazhong University of Science and Technology, Wuhan, China. His research focuses on agentic AI systems, safety of large/multi-modal language models, and AI for science (e.g., spatio-temporal data and biomedical applications). With over 40 publications in top-tier venues including ICLR, NeurIPS, ICML, KDD, AAAI, TPAMI, and TKDE, he has made significant contributions to the field. Dr. Wang was honored with the ICLR 2025 Best Paper Award and serves as a reviewer for leading journals and conferences such as TKDE, ICLR, TPAMI, NeurIPS, and ICML.
\end{minipage}\par\medskip

\clearpage

%% file: TriPhysBench_appendix.tex
\onecolumn
\shorttitle{TriPhysBench: Appendix}
\CitationPrefix{S}
\makeatletter\immediate\write\@auxout{\string\citation{BSTcontrol}}\makeatother
\appendix
\setcounter{table}{0}
\setcounter{figure}{0}
\setcounter{equation}{0}
\renewcommand{\thetable}{S\arabic{table}}
\renewcommand{\thefigure}{S\arabic{figure}}
\renewcommand{\theequation}{S\arabic{equation}}
\renewcommand{\thesubsection}{\thesection.\arabic{subsection}}
\renewcommand{\theHsection}{appendix.\Alph{section}}
\renewcommand{\theHsubsection}{\theHsection.\arabic{subsection}}
\renewcommand{\theHtable}{appendix.\arabic{table}}
\renewcommand{\theHfigure}{appendix.\arabic{figure}}
\renewcommand{\theHequation}{appendix.\arabic{equation}}
\pretocmd{\section}{\FloatBarrier\Needspace{8\baselineskip}}{}{\PackageError{triphys}{Appendix spacing patch failed}{Check the section definition.}}
\RecustomVerbatimEnvironment{verbatim}{Verbatim}{breaklines=true,breakanywhere=true,fontsize=\fontsize{8.5}{10.5}\selectfont,samepage=false,frame=single,framerule=0.3pt,framesep=5pt}
\suppressfloats[t]
\pdfbookmark[0]{Appendix}{appendix}

\begin{center}
{\Large\bfseries Appendix\par}
\vspace{4pt}
{\normalsize TriPhysBench: Evaluating Physical Intelligence for Embodied Agents\par}
\end{center}
\noindent
Appendices~\ref{app:construction}--\ref{app:results} provide benchmark specifications, evaluation details, the video-level partition, training procedures, and detailed results. Tables and references use the prefix S to distinguish them from the main paper.

\section{Benchmark Specification}\label{app:construction}

\subsection{Records and Evidence Roles}\label{records-and-evidence-roles}

TriPhysBench studies physical intelligence through the coordination of physical perception, physical knowledge retrieval, and physical reasoning. It comprises two evaluation subtasks: Measurement-grounded Physical Reasoning (MPR), centered on supported video measurements and recovery of withheld observations, and Knowledge-grounded Physical Reasoning (KPR), emphasizing the acquisition and application of external physical specifications. MPR uses 228 questions from 90 EgoExo4D videos \cite{grauman2024egoexo4d}; KPR uses 182 questions from 110 videos in BAAI-DataCube's Physics-aware-videos collection \cite{baai2025physicsaware}. Both subtasks combine video observations and physical knowledge in quantitative problems, and both include all four answer forms.

Each question is linked to a source recording. EgoExo4D items retain a take identifier, while Physics-aware-videos items retain folder and video identifiers. A recording can support several physical questions. Its questions stay together in the training/test partition, even if they concern different event intervals.

The three capabilities describe the task demands. A question record consists of a video \(V_i\), a question \(q_i\), an answer specification \(\mathcal Y_i\), an observation record \(E_i^{(1)}\), a physical knowledge record \(E_i^{(3)}\), and a reference solution \(\tau_i^\star\):

\[
\mathcal B_i=(V_i,q_i,\mathcal Y_i,E_i^{(1)},E_i^{(3)},\tau_i^\star).
\]

The solver receives the video and question, including any stated observations; reference annotations support construction and evaluation. Observation records identify the entity, quantity, temporal support, unit, and measurement source. Knowledge records identify the relevant entity and conditions for applying a property or relation. Table \ref{tab:record-fields} defines the principal fields. A source entry can identify a video interval, an external retrieval pointer, or a numerical operation.

\begin{table}[pos=t]
\caption{Fields in TriPhysBench question records.}
\label{tab:record-fields}
\centering\footnotesize
\setlength{\tabcolsep}{3.5pt}
\renewcommand{\arraystretch}{1.18}
\begin{tabularx}{0.96\linewidth}{lX}
\toprule
\textbf{Field} & \textbf{Interpretation} \\
\midrule
\shortstack[l]{take\_uid\\folder\_id and video\_id} & Source-video grouping keys, depending on the source \\
question; question\_type; difficulty & Problem statement, answer form, and construction tier \\
physics\_category & Parent physical category and subcategory annotation \\
answer; answer\_range & Reference answer and an interval when applicable \\
annotated\_solution & Ordered reference-solution entries \\
sources & Observation times, external-property pointers, and calculation records \\
required\_external\_tools & Operations expected by the reference solution \\
\shortstack[l]{answerable\_without\_video\\answerable\_without\_tools} & Construction-time judgment labels \\
\shortstack[l]{is\_question\_understandable\\is\_answer\_correct\_and\_unique} & Quality-review labels \\
valid & Recorded acceptance decision \\
\bottomrule
\end{tabularx}
\end{table}

Video- and tool-dependence labels record judgments made during construction. They do not measure model behavior under video or tool removal. A quantity stated in the question can be used directly, and a reference retrieval record does not show whether an evaluated model performed retrieval. The quality checks below describe how the supporting records are assessed.

Examples of the four answer forms can be located in the question records. The point-estimation example is question 1 of EgoExo4D take \texttt{72cfc694-\allowbreak 5d6f-\allowbreak 42cc-\allowbreak 918a-\allowbreak eb1ec0086896} (MPR). The interval-estimation example is question 1 of Physics-aware-videos folder \texttt{a00054}, video \texttt{00136}; the physical-judgment example is question 2 of video \texttt{00145} in the same folder; and the scenario-comparison example is question 1 of folder \texttt{a00128}, video \texttt{00907} (KPR). Question indices are one-based within each take or video. These records include reference conditions such as the whole-hand approximation of wrist speed, a typical hand mass, a range of rolling-friction coefficients, a curve radius and static-friction coefficient, and the thermal conductivities of paper and steel. The conditions are assumptions used in the corresponding reference solutions.

\subsection{Screening and Temporal Representation}\label{screening-and-temporal-representation}

MPR recordings undergo an initial screening by physical events. Candidates are EgoExo4D takes with three-dimensional body-pose annotations that last less than 60 s. Keyword lists for 15 physical event types, including projectile motion, collision, rotation, oscillation, friction, deformation, pouring, and free fall, are matched against the atomic action descriptions of each take. Takes with more detected event types are prioritized.

KPR recordings undergo an initial screening against the event categories of the physical-event library. The caption of every Physics-aware-videos recording is matched against keyword lists for these categories, and scene types without a measurable subject, such as ocean waves, cooking, and team sports, are excluded. The screened recordings are then inspected manually for technical quality, visible motion, and an identifiable subject that supports a quantitative question. The benchmark retains questions from 90 MPR takes and 110 KPR recordings, which range from 360p to 2160p and from 5.0 s to 60.0 s.

Gemini-3.6-Flash \cite{google2026gemini36flash} produces clip-level descriptions and visible-element lists. The temporal parsing procedure uses intervals of at least four seconds. The output contains a global summary and entities or events visible within the interval. Local times are associated with the original video so that a source entry can identify the temporal support for a quantity.

\subsection{Measurement Conditions and Question Generation}\label{measurement-conditions-and-question-generation}

For MPR, available three-dimensional EgoExo4D pose annotations provide reference geometry for eligible measurements. The upright-frame test forms a vector from the midpoint of the hips to the midpoint of the shoulders and measures its angle to the annotated vertical direction. Frames with an angle below 20 degrees are eligible for the upright reference; a clip without eligible support does not provide a body-height measurement through this route.

The human branch uses EgoExo4D three-dimensional body-pose annotations when available, and otherwise YOLO11-pose keypoints \cite{jocher2024yolo11} with MediaPipe pose estimates \cite{googlemediapipepose} as a fallback. The object branch links YOLO11n detections \cite{jocher2024yolo11} with ByteTrack \cite{zhang2022bytetrack}, and YOLO-World \cite{cheng2024yoloworld} locates reference objects of known size for scale calibration. Records retain the extraction method, confidence metadata, and temporal support. Each measurement is assigned a reliability tier according to its scale-calibration source, with relative tolerances of \(\pm5\%\), \(\pm12\%\), and \(\pm25\%\) for tiers A, B, and C; tier D is discarded. A measurement enters question construction only when its measurement confidence is at least 0.80 and its calibration confidence is at least 0.75, and repeated measurements of the same subject and quantity retain the higher-confidence record. These relative tolerances are construction tolerances, not confidence intervals estimated from a held-out calibration set. Metric interpretation also requires compatible camera geometry, coordinates, and timestamps.

For a tracked position \(\mathbf p(t)\) expressed in a common metric coordinate system, a finite-difference speed estimate is

\[
\widehat v_{[t,t+\Delta t]}=\frac{\|\mathbf p(t+\Delta t)-\mathbf p(t)\|_2}{\Delta t}.
\]

The time difference comes from the aligned recording and annotation timestamps. Camera motion and coordinate changes must be resolved before interpreting the difference as physical displacement. A change in motion direction is obtained from the angle between the incoming and outgoing nonzero velocity vectors in the same frame of reference. These conditions determine whether the measurement can support a metric answer.

For targets such as kinetic energy, the quadratic dependence on speed can amplify the adopted input uncertainty.

For KPR, the visual estimator targets scale-free observations in Physics-aware-videos. Angle records identify a reference direction and time interval, count records identify repeated events and their occurrences, and ratio records identify the target and comparison reference. Image-plane direction changes retain that interpretation unless the viewpoint supports a scene-level angle. Relative displacements and speeds require a compatible reference and temporal interval.

Searchable objects are associated with external mass, dimension, or material specifications. Their applicability depends on the object's identity and the physical assumptions in the question. Records retain source URLs and whether a property is a point value or a range. Any additional absolute value explicitly provided in a question functions as a problem condition; the scale-free route does not independently establish its metric calibration.

DeepSeek-V3.2 \cite{deepseek2025v32} organizes the observations and properties into an entity- and event-centered graph. Event templates specify required observations, external properties, relations, and output quantities. Graph nodes retain identifiers and evidence references; edges state the relation and its supporting record. A single extension pass matches the graph events to a physical-event library and adds the external properties and withheld-observation nodes that the matched templates require. The graph is used to construct reference solutions and questions, rather than being required as the internal representation of an evaluated model.

The graph schema records node fields \texttt{id}, \texttt{content}, \texttt{type}, \texttt{modality}, and \texttt{source\_info}. Edge fields \texttt{source}, \texttt{target}, \texttt{relation}, and \texttt{evidence} connect a relation to its supporting input. The available relation labels include causes, enables, triggers, results\_in, elaborates, and follows. Template requirements and source records constrain the dependencies considered during extension.

Difficulty is controlled within each subtask. MPR Easy, Medium, and Hard questions withhold zero, one, and two designated observations, respectively, subject to the available measurements. KPR questions state the estimated observations and withhold one, two, and three external specification values. Descriptions of an object can remain visible while its numerical property is withheld. Numeric-leak checks compare question text against withheld values and their rounded forms.

MPR construction requires at least one available measurement for Easy and Medium questions and at least two for Hard questions. The generation instructions require that doubling a withheld observation change the derived numerical answer by at least 20\%; this finite perturbation is a generation requirement intended to ensure that the hidden quantity affects the requested result. Leak detection compares the numbers in the question text with each withheld value rounded to zero through four decimal places, and a flagged question is regenerated once. For KPR, the number of available specifications limits the eligible tier. Retrieved specifications take precedence over a conflicting visual estimate when they refer to the same applicable object property. The construction also prohibits creating an additional observation merely by scaling another estimated quantity, such as inferring arm span from estimated height.

The four answer forms are point estimation, interval estimation, physical judgment, and scenario comparison, mapped to the original TYPE\_A, TYPE\_B, TYPE\_C, and TYPE\_D labels in Table \ref{tab:answer-forms}. These labels describe the reference answer within either subtask. Source codes A and B in the original experiment records correspond to MPR and KPR, respectively, and are distinct from the TYPE labels. Appendix~\ref{app:evaluation} specifies which model predictions the evaluation accepts.

\begin{table}[pos=!t]
\caption{Answer forms and original TYPE labels.}
\label{tab:answer-forms}
\centering\footnotesize
\setlength{\tabcolsep}{3.5pt}
\renewcommand{\arraystretch}{1.18}
\begin{tabularx}{0.94\linewidth}{lXX}
\toprule
\textbf{Type} & \textbf{Target} & \textbf{Reference form} \\
\midrule
TYPE\_A & Point estimate of a physical quantity & Numerical value with unit \\
TYPE\_B & Quantity represented by an uncertainty or specification range & Closed interval with unit \\
TYPE\_C & Physical judgment supported quantitatively & Yes/no with a numerical justification \\
TYPE\_D & Comparison between scenarios or quantities & Direction and numerical magnitude \\
\bottomrule
\end{tabularx}
\end{table}

The final collection contains 200 videos and 410 questions. Its 94 physics subcategory labels are counted before display normalization. Only the parent label Fluid\_Dynamics is displayed as Fluid dynamics in the distribution tables below; stored questions, answers, and scores are unchanged.

The four answer forms contain 160 point-estimation, 158 interval-estimation, 20 physical-judgment, and 72 scenario-comparison questions. MPR contributes 63 of the 72 comparison questions, while KPR contributes 97 of the 160 point-estimation questions.

The main statistics figure counts each scene or physics term at most once per question. Reference-solution length is the number of entries in \texttt{annotated\_solution}, with means of 6.45 for MPR and 6.66 for KPR. Median lengths across Easy, Medium, and Hard are 5, 6, and 7 entries for MPR, and 6, 7, and 8 for KPR. The corresponding boxplots contain 82/77/69 MPR questions and 90/57/35 KPR questions. Difficulty percentages use all 410 questions. Boxes span the interquartile range, with a median line; whiskers reach the most extreme observations within 1.5 interquartile ranges of the box, and dots mark outliers.

\subsection{Grouped Physics Topics}\label{app:physics-topics}

Panel (d) of the main statistics figure groups the 94 original physics labels into eight display topics, retaining all 410 questions. Each original label maps to exactly one topic. The grouping follows the annotated physical concept, with mixed or broad labels checked against their questions and reference solutions. For example, \texttt{Energy\_Kinetic} and \texttt{Kinetic\_Energy} both enter Work \& energy; fluid-specific labels remain in Fluid dynamics even when they involve energy or momentum. The fluid-related label under Mechanics also enters Fluid dynamics, giving 32 questions in this display group. The original parent-category counts used for partitioning remain unchanged.

The complete mapping follows. Prefixes abbreviate the original parent labels: M = Mechanics, B = Biomechanics, F = Fluid\_Dynamics, T = Thermodynamics, E = Electromagnetism, and O = Optics. Suffixes retain their original spelling. The accompanying \texttt{physics\_topic\_mapping.csv} records the mapping and its rationale for every label.

\input{tables/physics_topic_mapping}

\subsection{Quality Checks and Their Scope}\label{quality-checks-and-their-scope}

Construction review assesses understandability, answer consistency, support for observations and properties, and the logic of the reference solution. Large language model (LLM) review is followed by deterministic checks of fields, allowed labels, answer placeholders, and interval validity. Acceptance and rejection decisions remain in the question records.

Semantic review labels six aspects: question understandability, correctness and uniqueness of the reference answer, answerability without the video, answerability without external tools, multi-hop reasoning quality, and overall quality. A question is retained only when it is understandable, has a correct and unique answer, requires both the video and external tools, and receives a rating of good or excellent for both reasoning quality and overall quality. Rule filtering then removes questions with missing source records, an invalid answer interval, or an ambiguous object reference.

The review-stage model call disables tool use, so its decisions do not establish independent execution of every reference calculation. Six KPR questions retain a non-HTTP placeholder in the annotated-URL field, although their primary URL fields contain HTTP addresses. This syntax check identifies incomplete pointers but does not assess whether linked content is accessible or supports the numerical value.

An additional review covers all 410 questions (228 MPR and 182 KPR). GPT-5.5 and Gemini-3.6-Flash each judge whether a question is understandable and whether its reference answer is correct and unique. The judges receive the question, reference answer, reference solution, and the video-measured and retrieved values used by the solution; the prompt states that these hidden values are part of the design and have unique values. The third rater is the DeepSeek-V3.2 construction review, which is positive for every retained question. All three judges agree on understandability for 409 questions and on answer correctness for 392 questions (95.6\%; Table \ref{tab:agreement}). Of the 18 questions with a negative judgment, 16 belong to the training partition. Ten concern a modeling simplification in the reference solution, four the uniqueness of a quantity, three a mismatch between the retrieved value and the value used in the solution, and one an internal contradiction.

\begin{table}[pos=!t]
\caption{Model-judge agreement on question quality over all 410 questions.}
\label{tab:agreement}
\centering\footnotesize
\setlength{\tabcolsep}{3.5pt}
\renewcommand{\arraystretch}{1.18}
\begin{tabularx}{0.92\linewidth}{Xrrr}
\toprule
\textbf{Statistic} & \textbf{MPR} & \textbf{KPR} & \textbf{All} \\
\midrule
Assessed questions & 228 & 182 & 410 \\
Understandability: agreement of all three judges & 227 (99.6\%) & 182 (100\%) & 409 (99.8\%) \\
Answer correctness: agreement of all three judges & 217 (95.2\%) & 175 (96.2\%) & 392 (95.6\%) \\
Answer correctness: negative by GPT-5.5 / Gemini-3.6-Flash / both & 5 / 9 / 3 & 4 / 5 / 2 & 9 / 14 / 5 \\
Answer correctness: GPT-5.5 vs.~Gemini-3.6-Flash agreement & 96.5\% & 97.3\% & 96.8\% \\
\bottomrule
\end{tabularx}
\end{table}

\section{Evaluation Details}\label{app:evaluation}

\subsection{Extraction and Unit Normalization}\label{extraction-and-unit-normalization}

The answer parser first searches for an \texttt{answer} block and extracts text after \texttt{Final\ answer:}. If the block is absent, it searches the response for a numerical expression with a unit. A missing closing tag or incomplete final sentence can trigger model-based evaluation, allowing some nonconforming responses to be scored. For multi-turn responses, the parser reads the last turn.

Numeric normalization recognizes scientific notation and common dash variants, then converts the prediction and reference to the International System of Units (SI) using Table \ref{tab:units}. The answer-form rule applies after normalization. The compared quantities must also have the same physical dimension.

\begin{table}[pos=!t]
\caption{Unit conversions used for answer normalization.}
\label{tab:units}
\centering\footnotesize
\setlength{\tabcolsep}{3.5pt}
\renewcommand{\arraystretch}{1.18}
\begin{tabularx}{0.80\linewidth}{lX}
\toprule
\textbf{SI target} & \textbf{Supported unit: multiplier} \\
\midrule
Meter & mm: 0.001; cm: 0.01; m: 1; km: 1000 \\
Kilogram & mg: 0.000001; g: 0.001; kg: 1 \\
Joule & mJ: 0.001; J: 1; kJ: 1000 \\
Newton & N: 1; kN: 1000 \\
Second & ms: 0.001; s: 1; min: 60; h: 3600 \\
Meter per second & cm/s: 0.01; m/s: 1; km/h: 0.277778 \\
\bottomrule
\end{tabularx}
\end{table}

\subsection{Answer Scoring}\label{answer-scoring}

We evaluate answer correctness by comparing the normalized final answer with its reference. The answer correctness score (ACC) is the mean of question-level scores in \([0,1]\), including partial credit. For point estimation, let \(p\) be the prediction and \(g\) the nonzero reference, with relative error \(\delta=|p-g|/|g|\). The scoring rule is

\[
A_A(p,g)=\begin{cases}
1,&\delta\leq0.02,\\
\exp[-8(\delta-0.02)],&0.02<\delta\leq0.30,\\
0,&\delta>0.30.
\end{cases}
\]

A failed numeric parse or zero-valued reference receives zero. The score gives full credit within 2\% relative error, decays exponentially from 2\% to 30\%, and falls to zero beyond the 30\% cutoff.

For interval estimation, let \(P=[l_p,u_p]\) be the predicted interval and \(G=[l_g,u_g]\) the reference interval. The score combines overlap and center calibration:

\[
A_B(P,G)=0.5\,\operatorname{IoU}(P,G)+0.5\,\operatorname{EQS}(P,G),
\]

Here, intersection-over-union (IoU) measures interval overlap, and EQS is the center-calibration term:
\[
\operatorname{IoU}(P,G)=\frac{\max\bigl(0,\min(u_p,u_g)-\max(l_p,l_g)\bigr)}{\max(u_p,u_g)-\min(l_p,l_g)},\qquad
\operatorname{EQS}(P,G)=\max\!\left(0,\,1-\frac{|c_p-c_g|}{w_g}\right),
\]
where \(c_p\) and \(c_g\) are the interval midpoints and \(w_g=u_g-l_g\) is the reference width. The IoU denominator is the length of the smallest interval covering both, which equals the union length whenever the intervals overlap. A predicted interval that lies entirely within \(G\) receives full credit. For a point prediction \(p\),
\[
A_B(p,G)=\begin{cases}
1,& l_g\leq p\leq u_g,\\
\tfrac12\bigl[e^{-5d/w_g}+e^{-5(d+w_g)/w_g}\bigr],&\text{otherwise},
\end{cases}
\]
where \(d\) is the distance from \(p\) to the nearest boundary of \(G\). An outside point therefore receives about 0.50 next to the boundary, and its score decays exponentially with distance measured in reference widths. The range in the final answer is scored, both endpoints follow the same unit normalization, and a final answer without a parsable number receives zero. All 158 reference intervals have positive width. Table~\ref{tab:interval-examples} lists scores for the interval-estimation example.

\begin{table}[pos=!t]
\caption{Interval-estimation scores for the reference interval 8.54--12.26 mJ.}
\label{tab:interval-examples}
\centering\footnotesize
\setlength{\tabcolsep}{3.5pt}
\renewcommand{\arraystretch}{1.18}
\begin{tabularx}{0.80\linewidth}{Xrrr}
\toprule
\textbf{Prediction (mJ)} & \textbf{IoU} & \textbf{EQS} & \textbf{Score} \\
\midrule
Interval 9--12 & -- & -- & 1.000 \\
Interval 10--14 & 0.414 & 0.570 & 0.492 \\
Interval 7--9 & 0.087 & 0.355 & 0.221 \\
Interval 13--15 & 0 & 0.032 & 0.016 \\
Point 10.0 & -- & -- & 1.000 \\
Point 12.27 & -- & -- & 0.497 \\
Point 8.0 & -- & -- & 0.244 \\
Point 13.0 & -- & -- & 0.186 \\
Point 15.0 & -- & -- & 0.013 \\
\bottomrule
\end{tabularx}
\end{table}

For physical judgment, the extracted decision is normalized to yes/no. Positive forms include yes, true, and correct; negative forms include no, false, and incorrect. The answer score compares this decision with the reference. Quantitative justification is assessed by the solution rubric.

For scenario comparison, direction and magnitude each contribute half of the answer score:

\[
A_D=0.5\,\mathbf{1}[\widehat d=d^\star]+0.5\,A_A(\widehat m,m^\star).
\]

Here, \(d\) is the requested greater-than, less-than, or equality relation, and \(m\) is the numerical magnitude, such as a ratio or difference, specified by the question and reference answer.

\subsection{Solver Prompt and Generation}\label{solver-prompt-and-generation}

All evaluated models receive the following system instruction, followed by the video, question, and a type-specific answer instruction from Table \ref{tab:type-prompts}.

\par\addvspace{\medskipamount}\noindent
\begin{minipage}{\linewidth}
\begin{verbatim}
You are an expert multimodal AI assistant specializing in physics reasoning
and quantitative analysis of video content. You should think step by step to
answer the question. You may use available tools to assist with your analysis
if needed.

IMPORTANT REASONING CONSTRAINTS:
- When you make an assumption or estimate, commit to it and proceed
- You may refine an assumption ONCE if initial reasoning reveals an issue
- Do NOT repeatedly question or revise the same assumption more than twice
- Avoid circular reasoning loops that rehash uncertainties
- Move forward decisively once you have reasonable estimates

CRITICAL: You MUST structure your response in this exact format:

<answer>
Reasoning chain:
Step 1: [first reasoning step]
Step 2: [second reasoning step]
...

Final answer: [your answer]
</answer>
\end{verbatim}
\end{minipage}\par\medskip

\begin{table}[pos=!t]
\caption{Answer-form instructions in the solver prompt.}
\label{tab:type-prompts}
\centering\footnotesize
\setlength{\tabcolsep}{3.5pt}
\renewcommand{\arraystretch}{1.18}
\begin{tabularx}{0.92\linewidth}{lX}
\toprule
\textbf{Type} & \textbf{Instruction} \\
\midrule
TYPE\_A & Provide a numerical value with appropriate unit \\
TYPE\_B & Provide either a single value OR a range \\
TYPE\_C & Start with `Yes' or `No', followed by quantitative justification \\
TYPE\_D & Specify the comparison direction and magnitude \\
\bottomrule
\end{tabularx}
\end{table}

The evaluated models are GPT-5.5 \cite{openai2026gpt55}, Gemini-3.6-Flash \cite{google2026gemini36flash}, Qwen3.6-27B \cite{qwen2026qwen36_27b}, Qwen3.5-9B \cite{qwen2026qwen35_9b}, Qwen3.5-4B \cite{qwen2026qwen35_4b}, and InteractiveOmni-4B \cite{sensenova2025interactive}. Table \ref{tab:protocol} lists their inference settings. The Qwen backbones and the TriPhys models share one configuration, so the comparisons in Tables \ref{tab:training-results} and \ref{tab:stages} differ only in the model weights. In each turn, a model either calls tools or gives an answer block, and tool results are appended as tool messages. The environment record lists eight NVIDIA A800-SXM4-80GB GPUs, an Intel Xeon Platinum 8358 CPU, and 1 TB RAM, together with Python 3.10.21, PyTorch 2.5.1 \cite{paszke2019pytorch}, CUDA 12.1, Transformers 5.17.0 \cite{wolf2020transformers}, PEFT 0.20.0, and flash-linear-attention 0.3.2. Per-model adapters, checkpoint revisions, and service dates are unspecified.

\begin{table}[pos=!t]
\caption{Inference settings of the evaluated models.}
\label{tab:protocol}
\centering\footnotesize
\setlength{\tabcolsep}{3.5pt}
\renewcommand{\arraystretch}{1.18}
\begin{tabularx}{\linewidth}{lX}
\toprule
\textbf{Setting} & \textbf{Value or rule} \\
\midrule
\multicolumn{2}{l}{\textit{Qwen3.5-4B, Qwen3.5-9B, Qwen3.6-27B, and TriPhys models}} \\
Video input & Qwen-VL processor at 1 fps \\
Prompt & System instruction above; answer-form instruction from Table \ref{tab:type-prompts} \\
Tools & \texttt{web\_search}, \texttt{code\_executor}; declared through the chat template \\
Turns and budget & At most 6 turns of at most 8,192 tokens \\
Decoding & Temperature 0.1, top-p 0.9, top-k 20, repetition penalty 1.05 \\
\midrule
\multicolumn{2}{l}{\textit{GPT-5.5 and Gemini-3.6-Flash}} \\
Video input & GPT-5.5: frames at 2 fps, 8--48 per clip, longer side 768 pixels, each preceded by its timestamp; Gemini-3.6-Flash: whole clip as H.264 video, 10 fps at 480p for MPR and native frame rate at 720p for KPR \\
Prompt & Same as the Qwen models \\
Tools & Same tools; declared through function calling \\
Turns and budget & At most 6 turns of at most 8,192 output tokens \\
\midrule
\multicolumn{2}{l}{\textit{InteractiveOmni-4B}} \\
Video input & Official chat interface; 8 uniformly sampled frames, at most 12 image tiles per frame \\
Prompt & Same as the Qwen models \\
Tools & tools declared in the system message in the Qwen3 format \\
Turns and budget & At most 6 turns of at most 8,192 tokens \\
Decoding & Temperature 0.1, top-p 0.9 \\
\bottomrule
\end{tabularx}
\end{table}

\subsection{Solution Judge and Rubric}\label{solution-judge-and-rubric}

We assess the supporting derivation with a DeepSeek-V4-Flash \cite{deepseek2026v4flash} judge that compares the model solution with the reference. The judge receives only the reference solution and the model solution, without the question or video. The judge returns four component ratings and may include an explanation. Its configuration specifies temperature 0.3 and a 60-second request timeout with retry handling. The solution correctness score (SCS) combines the ratings as

\[
\mathrm{SCS}=0.25S_{\mathrm{obs}}+0.25S_{\mathrm{formula}}+0.30S_{\mathrm{derive}}+0.20S_{\mathrm{conclude}}.
\]

The judge receives the following prompt. Its observation criterion compares the quantities used in a solution, and their stated sources, with those of the reference solution. Reference solutions present quantities stated in the question as observation steps, so a solution that uses a given quantity with its stated value matches the reference in the same way as a quantity measured from the video, and retrieved properties are compared with the retrieved values of the reference.

\par\addvspace{\medskipamount}\noindent
\begin{minipage}{\linewidth}
\begin{verbatim}
You are a physics reasoning evaluation expert. Compare the student's reasoning
chain with the standard reasoning chain (Ground Truth), and evaluate the quality
of the student's reasoning chain.

Please score the student's reasoning chain across four dimensions using
five-level scoring (0, 0.25, 0.5, 0.75, 1.0):

1. **S_obs (Physical Quantity Observation Quality)**:
   - 1.0: Correctly identified and extracted all key physical quantities
     (velocity, mass, height, angle, etc.) from the video, with clear
     sources, consistent with GT
   - 0.75: Identified main quantities, few missing or unclear sources
   - 0.5: Only identified partial key quantities, or sources are questionable
   - 0.25: Only a few quantities mentioned, lacking video observation basis
   - 0.0: No valid physical quantities extracted from video

2. **S_formula (Physical Formula Correctness)**:
   - 1.0: Formula completely correct, applicable to the problem, consistent
     with GT formula
   - 0.75: Formula basically correct, minor notation or expression issues
   - 0.5: Formula partially correct, missing key terms
   - 0.25: Formula incorrect but reasoning direction somewhat reasonable
   - 0.0: No formula used or formula completely wrong

3. **S_derive (Derivation Step Validity)**:
   - 1.0: Derivation steps complete, logically rigorous, calculations
     correct, covering the main GT derivation steps
   - 0.75: Derivation basically complete, minor skips or calculation errors
   - 0.5: Derivation has obvious skips, but main logic correct
   - 0.25: Derivation missing many steps, logic chain unclear
   - 0.0: No derivation process or derivation completely wrong

4. **S_conclude (Conclusion Logic)**:
   - 1.0: Conclusion clear, consistent with derivation, matches GT conclusion
   - 0.75: Conclusion basically correct, slight expression flaws
   - 0.5: Conclusion partially correct, not sufficiently clear
   - 0.25: Conclusion unclear or inconsistent with derivation
   - 0.0: No conclusion or conclusion completely wrong
\end{verbatim}
\end{minipage}\par\medskip

The failure analysis uses component ratings of at most 0.5 to identify failures in variable identification, physical relation selection, numerical derivation, and conclusion formation. For question \(i\) and component \(k\), the failure indicator and subtask-level incidence are

\[
F_{ik}=\mathbf{1}[S_{ik}\le0.5],\qquad
R_k=\frac{1}{N}\sum_{i=1}^{N}F_{ik}.
\]

The four categories correspond to the observation, formula, derivation, and conclusion components, respectively, also labeled variable identification, relation selection, numerical derivation, and conclusion formation in the diagnostic counts. MPR and KPR have \(N=228\) and \(N=182\) questions, respectively, including those with correct final answers. Categories can overlap within a question, so their rates need not sum to one. These rubric-based indicators locate weaknesses in the returned solution without establishing their causal order.

\subsection{Score Aggregation and Reporting}\label{score-aggregation-and-reporting}

The result tables distinguish evaluation scope from scoring metric. MPR and KPR identify the two subtasks; Overall pools their question-level scores. Overall ACC and SCS are means across all questions in the relevant partition. The subtasks contain 228 and 182 questions in the full benchmark and 61 and 49 in testing.

The overall score (OGS) is \(0.70\,\mathrm{ACC}+0.30\,\mathrm{SCS}\) within the indicated evaluation scope. Full-benchmark tables retain the recorded four-decimal aggregates, without reconstructing these entries from the rounded values in other columns. For the training and stage comparisons, OGS and increments use the four-decimal ACC and SCS means, with decimal half-up rounding applied after calculation. Difficulty-subgroup OGS retains four-decimal precision. These descriptive aggregates provide no confidence intervals, significance tests, or estimates of variation across runs.

\section{Video-Level Partition}\label{app:partition}

\subsection{Assignment Procedure}\label{assignment-procedure}

Partitioning is performed separately for MPR and KPR at the level of source videos, so that all questions from one recording fall on the same side. Each video is characterized by its modal question type, difficulty, and parent physics category, and these attributes define the strata. Sparse strata are merged with related strata, and videos are allocated toward a 70:30 training/test ratio while keeping the question-level distributions of type, difficulty, and category close to those of the full pool.

This procedure preserves source-video groups, while stratification approximates the original distribution. Modal attributes cannot preserve the complete mixture of question labels within every video. Sparse labels and integer video counts also prevent exact 70:30 allocation. The realized split contains 66/24 training/test videos for MPR and 81/29 for KPR. MPR contributes 167 training and 61 test questions, and KPR contributes 133 and 49. Overall, the training/test ratios are 73.5:26.5 for videos and approximately 73.2:26.8 for questions. The training questions from both subtasks are pooled for adaptation.

The stored partition has no overlapping video keys, and its training and test questions jointly match each complete cleaned source collection. These checks establish separation at the source-identifier level, without establishing absence from model pretraining or independence of related scenes under distinct identifiers.

\subsection{Realized Distributions}\label{realized-distributions}

Question counts in Table \ref{tab:split-distribution} give the denominators for aggregate and subgroup comparisons.

\begin{table}[pos=!t]
\caption{Training and test distributions by subtask.}
\label{tab:split-distribution}
\centering\footnotesize
\setlength{\tabcolsep}{3.5pt}
\renewcommand{\arraystretch}{1.18}
\begin{tabularx}{0.94\linewidth}{lXrrrr}
\toprule
\textbf{Dimension} & \textbf{Category} & \textbf{MPR train} & \textbf{MPR test} & \textbf{KPR train} & \textbf{KPR test} \\
\midrule
Question type & TYPE\_A & 44 & 19 & 70 & 27 \\
Question type & TYPE\_B & 73 & 23 & 46 & 16 \\
Question type & TYPE\_C & 5 & 1 & 11 & 3 \\
Question type & TYPE\_D & 45 & 18 & 6 & 3 \\
\midrule
Difficulty & Easy & 58 & 24 & 65 & 25 \\
Difficulty & Medium & 56 & 21 & 42 & 15 \\
Difficulty & Hard & 53 & 16 & 26 & 9 \\
\midrule
Physics & Mechanics & 166 & 61 & 88 & 32 \\
Physics & Fluid dynamics & 0 & 0 & 22 & 9 \\
Physics & Thermodynamics & 0 & 0 & 19 & 6 \\
Physics & Electromagnetism & 0 & 0 & 3 & 2 \\
Physics & Optics & 0 & 0 & 1 & 0 \\
Physics & Biomechanics & 1 & 0 & 0 & 0 \\
\bottomrule
\end{tabularx}
\end{table}

The test subset contains 46 point-estimation, 39 interval-estimation, four physical-judgment, and 21 scenario-comparison questions. Difficulty counts are 49 Easy, 36 Medium, and 25 Hard. Each labeling dimension accounts for all 110 questions. The MPR test split contains only one physical-judgment question, and the KPR test split contains no optics questions. These sparse or absent categories limit subgroup comparisons.

\section{Trajectories and Training}\label{app:training}

TriPhys trains an agent to identify required observations, acquire missing physical knowledge, and use computation to derive an answer. It uses supervised fine-tuning (SFT) on full tool-integrated trajectories, followed by full-trajectory Direct Preference Optimization (DPO) \cite{rafailov2023dpo}. Both stages use the training partition defined in Appendix~\ref{app:partition}. The resulting TriPhys-4B, TriPhys-9B, and TriPhys-27B models are initialized from Qwen3.5-4B, Qwen3.5-9B, and Qwen3.6-27B, respectively. 

\subsection{Trajectory Sampling and Verification}\label{trajectory-sampling-and-verification}

The starting set comprises 300 training questions. The base model solves each question with tools, without access to the reference answer, and samples four trajectories per question. A trajectory passes verification when its final answer reaches an ACC of at least 0.9, its last turn contains an answer, it calls at least one tool, and no retrieval fails. For point estimates, this threshold corresponds to a relative error of about 3.3\% or less. For each question, the three verified trajectories with the highest SCS are kept.

Gemini-3.6-Flash rewrites only the final answer block of each retained trajectory into a stepwise chain, using text alone. DeepSeek-V4-Flash scores the rewrite with the evaluation SCS rubric, and a rewrite is accepted when its SCS is at least \(\max(0.75,\mathrm{SCS}_{0})\), with at most three attempts. Gemini-3.6-Flash also solves questions that the backbone never answers correctly, without access to the reference answer. Stalled samples give no answer, call tools in at least two turns, and have no retrieval failure; an intermediate SFT model continues their first two turns for at most three further turns, and continuations with ACC of at least 0.9 are retained. The prefix and continuation therefore come from different models.

Table \ref{tab:stage1-data} gives their sources. Multiple verified trajectories for the same question differ in query wording, tool order, and reasoning text, while the retrieved value and final answer remain unchanged. They broaden the expression of existing problems without adding new videos or physical scenarios.

\begin{table}[pos=!t]
\caption{Composition of Stage-I training trajectories.}
\label{tab:stage1-data}
\centering\footnotesize
\setlength{\tabcolsep}{3.5pt}
\renewcommand{\arraystretch}{1.18}
\begin{tabularx}{0.96\linewidth}{lXrrr}
\toprule
\textbf{Source} & \textbf{Construction} \\
\midrule
Self-sampled & Base-model sampling, verification, and answer-block rewriting \\
Teacher-solved & Gemini-3.6-Flash solving without the reference answer \\
Continued & Continuation of stalled samples by an intermediate SFT model \\
\bottomrule
\end{tabularx}
\end{table}

\subsection{A Linked Training Annotation}\label{a-linked-training-annotation}

The first question in EgoExo4D training take \texttt{72cfc694-\allowbreak 5d6f-\allowbreak 42cc-\allowbreak 918a-\allowbreak eb1ec0086896} asks:

\begin{quote}
In the video, a bicycle mechanic's wrist reaches a speed of 4.6115 m/s at approximately t=4.87 s while working on a front wheel assembly. What is the kinetic energy of the hand during this motion? Assume the entire hand moves with the wrist at that speed and its rotational kinetic energy is negligible.
\end{quote}

The reference annotation retrieves a hand mass of 409.6 g and applies the point-mass kinetic-energy relation. Under those values,

\[
E=\tfrac12mv^2=0.5\times0.4096\times(4.6115)^2=4.35526292\ \mathrm{J},
\]

which rounds to the stored answer, 4.36 J. The retrieved value is a mean from a published hand-volume study rather than a measurement of the participant. The question states the translational approximation that applies the wrist speed to the whole hand. The calculation reproduces the annotation under those assumptions.

The following English rendering illustrates the tool-integrated reasoning (TIR) format of a verified Stage-I trajectory for this question, with its rewritten answer block. It is a serialization example, not a candidate rollout observed from an evaluated model.

\par\addvspace{\medskipamount}\noindent
\begin{minipage}{\linewidth}
\begin{verbatim}
<think>Compute kinetic energy using KE = 0.5*m*v^2. Obtain the hand-mass property.</think>
<tool_call>{"name": "web_search",
  "arguments": {"query": "typical mass of a human hand"}}</tool_call>
<tool_response>... a weight of 409.6 g ...</tool_response>
<think>Use m = 0.4096 kg and the stated wrist speed v = 4.6115 m/s.</think>
<tool_call>{"name": "code_executor",
  "arguments": {"code": "print(0.5*0.4096*4.6115**2)"}}</tool_call>
<tool_response>4.35526292...</tool_response>
<answer>
Reasoning chain:
Step 1: Observation: the wrist speed is v = 4.6115 m/s at t = 4.87 s (given).
Step 2: Retrieved property: hand mass m = 409.6 g = 0.4096 kg.
Step 3: Formula: KE = 0.5*m*v^2 (translational; the whole hand moves with the wrist).
Step 4: Substitution: KE = 0.5*0.4096*4.6115^2 = 4.355 J.
Step 5: Conclusion: the kinetic energy of the hand is about 4.36 J.
Final answer: 4.36 J
</answer>
\end{verbatim}
\end{minipage}\par\medskip

The reasoning, tool calls, and tool returns retain their sampled form. Only the answer block is rewritten, and its final value must match that of the sampled trajectory.

\subsection{Loss Mask and Adaptation Settings}\label{loss-mask-and-adaptation-settings}

SFT predicts assistant-generated reasoning, tool-call, and rewritten answer tokens throughout each complete trajectory, and reasoning tokens are weighted equally with other assistant tokens. The system prompt, user input, padding, and external tool responses are excluded from the prediction loss. Tool observations remain in the context of later generated segments. Thus, loss masking does not remove the information returned by a tool from attention.

Low-rank adaptation (LoRA) supplies the trainable weights \cite{hu2021lora}, with its alpha-to-rank ratio scaling the low-rank update. The adapters contain 42.5M, 58.2M, and 159.4M trainable parameters for the 4B, 9B, and 27B backbones, or 0.93\%, 0.61\%, and 0.58\% of their parameters. Table \ref{tab:training-settings} gives the settings used for training. SFT uses a learning rate of \(1\times10^{-4}\) for three epochs, with 14 steps per epoch and 42 steps in total; the schedule uses a four-step linear warm-up followed by cosine decay. Both stages use a maximum sequence length of 16,384 tokens, and videos are sampled at 1 fps and uniformly downsampled when they exceed this budget. Seed 42 governs both data assignment and training, and each configuration is trained once.

\begin{table}[pos=!t]
\caption{Training settings for full-trajectory SFT and DPO.}
\label{tab:training-settings}
\centering\footnotesize
\setlength{\tabcolsep}{3.5pt}
\renewcommand{\arraystretch}{1.18}
\begin{tabularx}{0.96\linewidth}{lX}
\toprule
\textbf{Setting} & \textbf{Value or rule} \\
\midrule
Backbones & Qwen3.5-4B, Qwen3.5-9B, Qwen3.6-27B \\
Training data & 300 questions from 147 videos (167 MPR / 133 KPR) \\
Tools & Native tool format; \texttt{web\_search} (Serper, top 10 results), \texttt{code\_executor} (Python, 60 s timeout) \\
Stage-I sampling model & Base model with tools, without the reference answer \\
Stage-I sampling & 4 per question; temperature 0.7, top-p 0.9, top-k 20, repetition penalty 1.05; at most 6 turns of at most 8,192 tokens \\
Stage-I verification & ACC \(\ge\) 0.9, answer in the last turn, at least one tool call, no retrieval failure; top 3 per question by SCS; subtask balance \\
Answer-block rewriting & Gemini-3.6-Flash, answer block only; accepted when SCS \(\ge\max(0.75,\mathrm{SCS}_{0})\); at most 3 attempts \\
SFT loss targets & Assistant-generated tokens (reasoning, tool calls, answer block); tool returns used as context; reasoning-token weight 1.0 \\
Maximum sequence length & 16,384 tokens; video at 1 fps, uniformly downsampled to at most 40 frames \\
LoRA rank / alpha / dropout & 32 / 64 / 0.05 \\
LoRA projections & q\_proj, k\_proj, v\_proj, o\_proj, gate\_proj, up\_proj, down\_proj \\
SFT learning rate / epochs & \(1\times10^{-4}\) / 3 \\
SFT per-device / effective batch & 1 / 16 \\
SFT schedule & 14 steps per epoch, 42 in total; 4-step linear warm-up followed by cosine decay; AdamW, weight decay 0, gradient clipping 1.0 \\
Preference sampling model & SFT checkpoint of TriPhys \\
Preference sampling & At most 4 attempts and 2 prefixes per question, 4 complete candidates per prefix; temperature 0.7, top-p 0.9, top-k 20, repetition penalty 1.05 \\
Acceptance & Answered, ACC \(\ge\) 0.9, reference value used in code or answer, SCS \(\ge\) 0.75 \\
Rejection & Unanswered or ACC \(<\) 0.5; candidates with retrieval failures removed \\
Chosen / rejected & Highest-SCS accepted candidate, else highest-SCS Stage-I trajectory / wrong and unanswered candidates, alternating \\
Pair retention & At least one rejected candidate and a chosen solution; at most 2 pairs per question \\
Reference policy & Frozen SFT checkpoint \\
DPO initialization & Merge SFT LoRA; train a new LoRA adapter \\
DPO loss & Length-normalized DPO, \(\beta=0.1\) (\(\beta\bar n=265.1\)); shared prefix masked; subtask weights 1.153 / 0.883; no NLL term \\
DPO learning rate / epochs & \(5\times10^{-6}\) / 2 \\
DPO per-device / effective batch & 1 / 16 pairs \\
DPO schedule & 9 steps per epoch, 18 in total; 10\% linear warm-up followed by cosine decay; AdamW, weight decay 0, gradient clipping 1.0 \\
Training seed & 42; each configuration trained once \\
Evaluation & Same as the pretrained backbones (Table \ref{tab:protocol}): temperature 0.1, top-p 0.9, top-k 20, repetition penalty 1.05; at most 6 turns of at most 8,192 tokens \\
\bottomrule
\end{tabularx}
\end{table}

\subsection{Preference Selection and Rejected-Trajectory Labels}\label{preference-selection-and-rejected-trajectory-labels}

Each training input receives at most four sampling attempts. At most two prefixes that retrieve the reference value are kept, and each is continued into four complete candidates, counting the original turn as one. The binary filter accepts an answer with an ACC of at least 0.9, compared with the 2\% full-credit region and continuous decay in evaluation ACC, and rejects a candidate that gives no answer or has an ACC below 0.5. Candidates with \(0.5\le\mathrm{ACC}<0.9\) are excluded, so nearly correct solutions do not serve as rejected examples, and candidates with retrieval failures are removed.

A preference pair is retained when the rejected set is nonempty and a chosen solution exists, selecting the accepted candidate with the highest SCS or, when none exists, the Stage-I trajectory of the question with the highest SCS. The fallback lets these questions contribute comparisons between correct and incorrect solutions. Acceptance also requires the reference value in the code or answer and an SCS of at least 0.75, excluding candidates whose answer falls in range through another constant. DPO compares the complete generated sequences against the frozen SFT reference \cite{rafailov2023dpo}. The trajectories share only the system prompt, video, and question, which are masked, and selection does not use a first-error location or local repair. This rule combines outcome and solution-quality preferences without using length, and the length-normalized loss keeps a long rejected trajectory from dominating the margin through its token count.

Rejected trajectories receive one of two labels, wrong answer or no answer, and the rejected trajectories of each question alternate between them. The pairs therefore cover both the misuse of a retrieved value and repeated retrieval without an answer. These labels serve only for selection and do not enter a reward term or category-specific loss. The DPO schedule specifies two epochs with an effective batch of 16 pairs, giving nine steps per epoch and 18 updates.

\section{Detailed Results}\label{app:results}

\subsection{Full-Benchmark Results by Subtask}\label{full-benchmark-results-by-subtask}

Table \ref{tab:source-results} provides the aggregate answer, solution, and overall scores for each subtask. Appendix~\ref{app:evaluation} describes aggregation and reporting precision.

For the three diagnostic models, weighting the four-decimal difficulty-subgroup OGS values by 172 Easy, 134 Medium, and 104 Hard questions agrees with the overall scores at that precision. Difficulty-subgroup ACC and SCS values are unavailable.

\begin{table}[pos=!t]
\caption{Full-benchmark results by subtask.}
\label{tab:source-results}
\centering\footnotesize
\setlength{\tabcolsep}{3.5pt}
\renewcommand{\arraystretch}{1.18}
\begin{tabularx}{0.84\linewidth}{Xlrrr}
\toprule
\textbf{Model} & \textbf{Subtask} & \textbf{ACC} & \textbf{SCS} & \textbf{OGS} \\
\midrule
GPT-5.5 & MPR & 0.4316 & 0.6269 & 0.4902 \\
GPT-5.5 & KPR & 0.4597 & 0.7076 & 0.5341 \\
\addlinespace[2pt]
Gemini-3.6-Flash & MPR & 0.4267 & 0.6390 & 0.4904 \\
Gemini-3.6-Flash & KPR & 0.6004 & 0.7453 & 0.6439 \\
\addlinespace[2pt]
Qwen3.6-27B & MPR & 0.4144 & 0.5854 & 0.4657 \\
Qwen3.6-27B & KPR & 0.4672 & 0.7008 & 0.5373 \\
\addlinespace[2pt]
Qwen3.5-9B & MPR & 0.3924 & 0.5706 & 0.4459 \\
Qwen3.5-9B & KPR & 0.4536 & 0.6929 & 0.5254 \\
\addlinespace[2pt]
Qwen3.5-4B & MPR & 0.3739 & 0.5321 & 0.4214 \\
Qwen3.5-4B & KPR & 0.4151 & 0.6732 & 0.4925 \\
\addlinespace[2pt]
InteractiveOmni-4B & MPR & 0.3650 & 0.5142 & 0.4097 \\
InteractiveOmni-4B & KPR & 0.2770 & 0.5190 & 0.3496 \\
\bottomrule
\end{tabularx}
\end{table}

\subsection{Diagnostic Counts}\label{diagnostic-counts}

Table \ref{tab:failure-counts} gives the counts underlying panel (b) of the main failure-analysis figure. The columns correspond to failures in variable identification, physical relation selection, numerical derivation, and conclusion formation, following the rubric mapping in Appendix~\ref{app:evaluation}. Each count is divided by 228 for MPR and 182 for KPR to obtain its percentage; a question can have more than one failure type.

\begin{table}[pos=!t]
\caption{Component failure counts and percentages.}
\label{tab:failure-counts}
\centering\footnotesize
\setlength{\tabcolsep}{3.5pt}
\renewcommand{\arraystretch}{1.18}
\begin{tabularx}{0.98\linewidth}{Xlrrrr}
\toprule
\textbf{Model} & \textbf{Subtask} & \shortstack[r]{\textbf{Variable}\\\textbf{Identification}} & \shortstack[r]{\textbf{Relation}\\\textbf{Selection}} & \shortstack[r]{\textbf{Numerical}\\\textbf{Derivation}} & \shortstack[r]{\textbf{Conclusion}\\\textbf{Formation}} \\
\midrule
Gemini-3.6-Flash & MPR & 158 (69.3\%) & 16 (7.0\%) & 110 (48.2\%) & 149 (65.4\%) \\
Gemini-3.6-Flash & KPR & 102 (56.0\%) & 10 (5.5\%) & 31 (17.0\%) & 66 (36.3\%) \\
\addlinespace[2pt]
Qwen3.5-4B & MPR & 197 (86.4\%) & 37 (16.2\%) & 145 (63.6\%) & 179 (78.5\%) \\
Qwen3.5-4B & KPR & 106 (58.2\%) & 17 (9.3\%) & 66 (36.3\%) & 120 (65.9\%) \\
\addlinespace[2pt]
InteractiveOmni-4B & MPR & 209 (91.7\%) & 48 (21.1\%) & 159 (69.7\%) & 196 (86.0\%) \\
InteractiveOmni-4B & KPR & 159 (87.4\%) & 61 (33.5\%) & 121 (66.5\%) & 147 (80.8\%) \\
\bottomrule
\end{tabularx}
\end{table}

\subsection{Test-Subset Increments}\label{test-subset-increments}

All three TriPhys models improve on the 110-question test subset (Table \ref{tab:test-increments}). Each difference compares a trained model with its corresponding pretrained backbone on the same questions. The main article provides their absolute scores.

\begin{table}[pos=!htbp]
\caption{Test-set gains over the corresponding pretrained backbones.}
\label{tab:test-increments}
\centering\footnotesize
\setlength{\tabcolsep}{3.5pt}
\renewcommand{\arraystretch}{1.18}
\begin{tabularx}{0.72\linewidth}{Xrrr}
\toprule
\textbf{Model} & \textbf{Delta ACC} & \textbf{Delta SCS} & \textbf{Delta OGS} \\
\midrule
TriPhys-4B & +0.0780 & +0.0577 & +0.0719 \\
TriPhys-9B & +0.0768 & +0.0398 & +0.0657 \\
TriPhys-27B & +0.0877 & +0.0372 & +0.0726 \\
\bottomrule
\end{tabularx}
\end{table}

For the training-stage ablations, derived OGS values before rounding are 0.51261, 0.56140, and 0.58452 for the backbone, SFT, and DPO checkpoints of TriPhys-4B; 0.52684, 0.57571, and 0.59254 for TriPhys-9B; and 0.54939, 0.55761, and 0.62194 for TriPhys-27B. The successive increments are 0.04879 and 0.02312 for TriPhys-4B, 0.04887 and 0.01683 for TriPhys-9B, and 0.00822 and 0.06433 for TriPhys-27B, summing to 0.07191, 0.06570, and 0.07255; the SFT shares are 67.8\%, 74.4\%, and 11.3\%. Rounding these results only at the final step yields the displayed increments.

These increments describe the cumulative sequence. The experiment includes no DPO-only model or isolated answer-rewriting and masking ablations, and post-training diagnostic counts are unavailable. The results therefore quantify improvements between the observed stages without separating individual training components.

\begingroup
\patchcmd{\section}{\FloatBarrier}{}{}{}
\renewcommand{\refname}{Appendix References}
\bibliographystyle{IEEEtran}
\bibliography{references,LingPhysKnow_preview_additions}
\endgroup
\shorttitle{TriPhysBench: Appendix}

%% file: tables/physics_topic_mapping.tex
\begingroup\footnotesize\raggedright
\setlength{\parskip}{3pt}
\noindent\textbf{Work \& energy.} \texttt{B/\allowbreak{}Energy\_\allowbreak{}Distribution}, \texttt{M/\allowbreak{}Biomechanics}, \texttt{M/\allowbreak{}Conservation\_\allowbreak{}of\_\allowbreak{}Energy}, \texttt{M/\allowbreak{}Energy}, \texttt{M/\allowbreak{}Energy\_\allowbreak{}Comparison}, \texttt{M/\allowbreak{}Energy\_\allowbreak{}Conservation}, \texttt{M/\allowbreak{}Energy\_\allowbreak{}Conservation\_\allowbreak{}Collision\_\allowbreak{}Friction}, \texttt{M/\allowbreak{}Energy\_\allowbreak{}Conservation\_\allowbreak{}Inclined\_\allowbreak{}Plane}, \texttt{M/\allowbreak{}Energy\_\allowbreak{}Conservation\_\allowbreak{}Rotational\_\allowbreak{}Dynamics}, \texttt{M/\allowbreak{}Energy\_\allowbreak{}Conservation\_\allowbreak{}with\_\allowbreak{}Friction}, \texttt{M/\allowbreak{}Energy\_\allowbreak{}Conversion}, \texttt{M/\allowbreak{}Energy\_\allowbreak{}Kinematics}, \texttt{M/\allowbreak{}Energy\_\allowbreak{}Kinetic}, \texttt{M/\allowbreak{}Energy\_\allowbreak{}Momentum}, \texttt{M/\allowbreak{}Energy\_\allowbreak{}Momentum\_\allowbreak{}Transfer}, \texttt{M/\allowbreak{}Energy\_\allowbreak{}Transfer}, \texttt{M/\allowbreak{}Energy\_\allowbreak{}Transfer\_\allowbreak{}Impact}, \texttt{M/\allowbreak{}Energy\_\allowbreak{}Transformation}, \texttt{M/\allowbreak{}Gravitational\_\allowbreak{}Energy}, \texttt{M/\allowbreak{}Inclined\_\allowbreak{}Plane\_\allowbreak{}Rolling\_\allowbreak{}Friction}, \texttt{M/\allowbreak{}Kinetic\_\allowbreak{}Energy}, \texttt{M/\allowbreak{}Kinetic\_\allowbreak{}Energy\_\allowbreak{}Biomechanics}, \texttt{M/\allowbreak{}Kinetic\_\allowbreak{}Energy\_\allowbreak{}Comparison}, \texttt{M/\allowbreak{}Kinetic\_\allowbreak{}Energy\_\allowbreak{}Ratio}, \texttt{M/\allowbreak{}Kinetic\_\allowbreak{}Energy\_\allowbreak{}Transfer}, \texttt{M/\allowbreak{}Work-\allowbreak{}Energy\_\allowbreak{}Theorem}, \texttt{M/\allowbreak{}Work\_\allowbreak{}Energy}, \texttt{M/\allowbreak{}Work\_\allowbreak{}Energy\_\allowbreak{}Friction}, \texttt{M/\allowbreak{}Work\_\allowbreak{}Energy\_\allowbreak{}Power}, \texttt{M/\allowbreak{}Work\_\allowbreak{}Energy\_\allowbreak{}Theorem}.\par
\noindent\textbf{Motion \& trajectories.} \texttt{M/\allowbreak{}Free\_\allowbreak{}Fall}, \texttt{M/\allowbreak{}Inclined\_\allowbreak{}Plane}, \texttt{M/\allowbreak{}Inclined\_\allowbreak{}Plane\_\allowbreak{}FreeFall\_\allowbreak{}Impact}, \texttt{M/\allowbreak{}Projectile\_\allowbreak{}Motion}, \texttt{M/\allowbreak{}Projectile\_\allowbreak{}Motion\_\allowbreak{}Energy}, \texttt{M/\allowbreak{}Projectile\_\allowbreak{}Motion\_\allowbreak{}with\_\allowbreak{}Drag}.\par
\noindent\textbf{Rotation \& oscillation.} \texttt{M/\allowbreak{}Circular\_\allowbreak{}Motion}, \texttt{M/\allowbreak{}Inclined\_\allowbreak{}Plane\_\allowbreak{}Rolling\_\allowbreak{}Motion}, \texttt{M/\allowbreak{}Oscillations}, \texttt{M/\allowbreak{}Pendulum\_\allowbreak{}Oscillation}, \texttt{M/\allowbreak{}Rolling\_\allowbreak{}Friction\_\allowbreak{}Inclined\_\allowbreak{}Plane}, \texttt{M/\allowbreak{}Rolling\_\allowbreak{}Motion}, \texttt{M/\allowbreak{}Rolling\_\allowbreak{}Motion\_\allowbreak{}Energy}, \texttt{M/\allowbreak{}Rolling\_\allowbreak{}Motion\_\allowbreak{}Inclined\_\allowbreak{}Plane}, \texttt{M/\allowbreak{}Rolling\_\allowbreak{}Motion\_\allowbreak{}on\_\allowbreak{}Inclined\_\allowbreak{}Plane}, \texttt{M/\allowbreak{}Rotational\_\allowbreak{}Dynamics}, \texttt{M/\allowbreak{}Rotational\_\allowbreak{}Dynamics\_\allowbreak{}Energy\_\allowbreak{}Conversion}, \texttt{M/\allowbreak{}Rotational\_\allowbreak{}Kinematics}, \texttt{M/\allowbreak{}Rotational\_\allowbreak{}Motion}, \texttt{M/\allowbreak{}Rotational\_\allowbreak{}to\_\allowbreak{}Linear\_\allowbreak{}Motion}, \texttt{M/\allowbreak{}Spring\_\allowbreak{}Oscillation}.\par
\noindent\textbf{Forces \& momentum.} \texttt{M/\allowbreak{}Beam\_\allowbreak{}Bending}, \texttt{M/\allowbreak{}Collision}, \texttt{M/\allowbreak{}Collision\_\allowbreak{}Dynamics}, \texttt{M/\allowbreak{}Collision\_\allowbreak{}Energy\_\allowbreak{}Transfer}, \texttt{M/\allowbreak{}Friction}, \texttt{M/\allowbreak{}Friction\_\allowbreak{}Kinematics}, \texttt{M/\allowbreak{}Friction\_\allowbreak{}Torque}, \texttt{M/\allowbreak{}Granular\_\allowbreak{}Flow}, \texttt{M/\allowbreak{}Impulse-\allowbreak{}Momentum}, \texttt{M/\allowbreak{}Impulse\_\allowbreak{}Momentum}, \texttt{M/\allowbreak{}Kinematics\_\allowbreak{}Dynamics}, \texttt{M/\allowbreak{}Linear\_\allowbreak{}Momentum}, \texttt{M/\allowbreak{}Momentum\_\allowbreak{}Transfer}, \texttt{M/\allowbreak{}Pulley\_\allowbreak{}System}, \texttt{M/\allowbreak{}Sliding\_\allowbreak{}Friction}, \texttt{M/\allowbreak{}Statics}.\par
\noindent\textbf{Fluid dynamics.} \texttt{F/\allowbreak{}Boundary\_\allowbreak{}Layer\_\allowbreak{}Flow}, \texttt{F/\allowbreak{}Buoyancy}, \texttt{F/\allowbreak{}Buoyancy\_\allowbreak{}Floating}, \texttt{F/\allowbreak{}Droplet\_\allowbreak{}Impact}, \texttt{F/\allowbreak{}Fluid\_\allowbreak{}Pressure}, \texttt{F/\allowbreak{}Impulse\_\allowbreak{}Momentum}, \texttt{F/\allowbreak{}Jet\_\allowbreak{}Impact}, \texttt{F/\allowbreak{}Pouring\_\allowbreak{}Flow}, \texttt{F/\allowbreak{}Pressure\_\allowbreak{}Difference}, \texttt{F/\allowbreak{}Splashing\_\allowbreak{}Impact}, \texttt{F/\allowbreak{}Surface\_\allowbreak{}Tension\_\allowbreak{}Instability}, \texttt{F/\allowbreak{}Terminal\_\allowbreak{}Velocity}, \texttt{F/\allowbreak{}Viscous\_\allowbreak{}Flow}, \texttt{M/\allowbreak{}Fluid\_\allowbreak{}Buoyancy}.\par
\noindent\textbf{Thermal physics.} \texttt{T/\allowbreak{}Combustion}, \texttt{T/\allowbreak{}Convection}, \texttt{T/\allowbreak{}Evaporation\_\allowbreak{}Condensation}, \texttt{T/\allowbreak{}Frictional\_\allowbreak{}Heating}, \texttt{T/\allowbreak{}Heat\_\allowbreak{}Transfer}, \texttt{T/\allowbreak{}Melting\_\allowbreak{}Solidification}, \texttt{T/\allowbreak{}Phase\_\allowbreak{}Change}, \texttt{T/\allowbreak{}Phase\_\allowbreak{}Change\_\allowbreak{}Heat\_\allowbreak{}Transfer}, \texttt{T/\allowbreak{}Phase\_\allowbreak{}Transition}, \texttt{T/\allowbreak{}Radiation}.\par
\noindent\textbf{Electromagnetism.} \texttt{E/\allowbreak{}Electrostatic\_\allowbreak{}Discharge}, \texttt{E/\allowbreak{}Magnetic\_\allowbreak{}Attraction\_\allowbreak{}Repulsion}.\par
\noindent\textbf{Optics.} \texttt{O/\allowbreak{}Shadow\_\allowbreak{}Formation}.\par
\endgroup